\documentclass[12pt]{article}

\usepackage[centertags]{amsmath}
\usepackage{latexsym,enumerate,subcaption}
\usepackage{amsmath,amsthm,amsopn,amstext,amscd,amsfonts,amssymb}
\usepackage{natbib}
\usepackage[dvips]{graphicx}
\usepackage{color}
\usepackage{comment}
\usepackage{tcolorbox}
\usepackage{multirow} 
\usepackage{booktabs}
\usepackage{url}
\RequirePackage[misc]{ifsym}
\usepackage{float}
\usepackage{hyperref}
\usepackage{listings}
\hypersetup{
    colorlinks=true,
    linkcolor=blue,          
    citecolor=blue,        
    filecolor=blue,         
    urlcolor=blue,
   }
\usepackage{enumitem}
\usepackage[toc,page]{appendix}
\usepackage{titlesec}  
\usepackage{diagbox}
\usepackage{orcidlink}

\date{}
\begin{document}
\title{\bf Processing and classifying bird songs using wavelet techniques and supervised learning}
\author{}
\maketitle
\vspace{-2cm}
\begin{center}
\author{Laura Lucia Dominguez Barrios}\orcidlink{0000-0001-6235-4806}\footnote{\Letter: \text{l175089@dac.unicamp.br}}
\author{Fidel Aniano Causil Barrios}
\orcidlink{0009-0008-7241-4655}\footnote{\Letter: \text{f244960@dac.unicamp.br}}
\author{Alex Rodrigo dos Santos Sousa}\orcidlink{0000-0001-5887-3638}\footnote{\Letter: \text{asousa@unicamp.br}}
\author{Mariana Rodrigues Motta}\orcidlink{0000-0002-2657-3857}\footnote{\Letter: \text{marirm@unicamp.br}}\\
{\footnotesize
$^{1,2,3,4}$Department of Statistics, State University of Campinas, Campinas, Brazil}
\end{center}

\maketitle

\section*{Abstract}
This study proposes an integrated framework for the processing and classification of invasive bird species vocalizations within natural soundscapes, characterized by high levels of environmental noise. We address the challenge of signal degradation by employing a Bayesian wavelet shrinkage methodology based on the Epanechnikov kernel prior, which offers a closed form decision rule and high computational efficiency for processing large bioacoustic datasets. The methodology was applied to recordings of three species obtained from the iNaturalist platform: \textit{Euphonia violacea}, \textit{Leiothrix lutea}, and \textit{Passer domesticus}. After signal denoising, we extracted a comprehensive set of features, including Mel-Frequency Cepstral Coefficients (MFCCs) and spectral indices such as entropy and zero-crossing rate. Several supervised learning models: Random Forest, Multinomial Logistic Regression and Support Vector Machine (SVM) were evaluated across different feature dimensionalities. Our results demonstrate that the proposed wavelet based preprocessing significantly enhances classification performance, with the SVM model achieving the highest accuracy (up to 0.9398) under a 10-dimensional MFCC configuration. This research provides a robust statistical tool for automated ecological monitoring and the management of biological invasions.

\paragraph{Keywords}  Bioacoustics; wavelet; bayesian; classification; shrinkage; invasive species; supervised learning.

\section{Introduction}
Animal acoustic signals are shaped by selection to convey information based on their tempo, intensity, and frequency. However, sound signals degrade as they transmit over space and across physical obstacles (e.g., vegetation or infrastructure), which affects communication potential. Therefore, propagation experiments are designed to quantify changes in signal structure in a given habitat by broadcasting and re-recording animal sounds at increasing distances \citep{baRulho}.\\

\noindent The extinction of species presents an irreversible loss to humanity, and preventing biodiversity loss is one of the biggest challenges our society faces. There are challenges to sound-based recognition of individual species in any natural soundscape, such as due to a variable distance of the sound source (the animal) to the sensor (microphone); inter-species and inter-individual variation in vocalizations; multiple unique vocalizations (sonotypes) per species, including mimicry; or biases due to equipment \citep{problems}. \\

\noindent Wavelet-based methods are widely applied in a range of fields, such as mathematics,signal and image processing, geophysics, and many others. In statistics, applications of wavelets arise mainly in the areas of non-parametric regression, density estimation,functional data analysis and stochastic processes. These methods basically utilize the possibility of representing functions that belong to certain functional spaces as expansions in wavelet basis, similar to other expansions such as polynomials, splines and Fourier. In particular, wavelet basis expansions are attractive due to their sparsity and localization properties; that is, most wavelet coefficients are equal to zero, while the few significant coefficients are concentrated around important features of the function, such as peaks, discontinuities, and oscillations. See \cite{daubechies1992} and \cite{mallat2009} for the theoretical background on wavelets, and \cite{vidakovic1999} and \cite{nason2008} for general statistical modeling using wavelets.  For further applications of wavelet-based methods to birdsong analysis and classification, see \cite{selin2006wavelets}, \cite{priyadarshani2016birdsong}, \cite{hsu2018local}, \cite{priyadarshani2020wavelet}, and, more recently, \cite{li2025multi}.
\\

\noindent Bayesian approaches have also been extensively explored because they allow the incorporation of prior information, such as the sparsity property and the support of the coefficients,
which is helpful in improving the estimates. The standard prior to the wavelet coefficients is
of spike and slab type. Several prior distributions for the wavelet coefficients have been proposed in the literature. Most of them are composed of a mixture of a point mass at zero and a symmetric unimodal distribution, such as the normal \citep{chipman-1997}, double-exponential \citep{vidakovic-ruggeri-bams}, double-Weibull \citep{remenyi-2015}, beta \citep{alex-beta} and logistic \citep{logistic} distributions. Although the Bayesian shrinkage rules available in the literature have been successfully applied to
several real data problems, they often do not perform well in data with high noise levels. Recently, \citep{fidel} proposed the use of a mixture of a point mass at zero and the Epanechnikov distribution as a prior for the wavelet coefficients, and the associated shrinkage rule showed good performance under high noise levels in simulation studies. \\


\noindent Our main objective is to develop a model for the classification of invasive birds based on bioacoustic signals obtained from natural soundscapes. These signals usually contain high levels of environmental noise from the recording sites, such as wind, insect sounds, and vocalizations from other species. Initially, the signals undergo a preprocessing stage in which silent time intervals are removed, the sampling rate (Hz) is standardized, and a wavelet shrinkage rule is applied for noise reduction and recovery of the true bird song.

\noindent After preprocessing, covariates strongly associated with the bioacoustic signals are extracted, such as Mel-Frequency Cepstral Coefficients (MFCC). These features are then used to train different supervised classification models available in the literature, allowing the subsequent evaluation and comparison of their performance in the identification of invasive birds.\\

\noindent \cite{classification} obtained results suggest that transfer learning and data augmentation could make the use of CNNs to classify species’ vocalizations feasible even for small soundscape-based projects with many rare species.\\

\noindent The remainder of this paper is organized as follows. Section 2 presents a brief overview of wavelet shrinkage. Section 3 describes the materials and methods, including the datasets and the supervised learning methods. Section 4 presents and discusses the results of the statistical analysis. Finally, Section 5 concludes the paper with final remarks. 

\section{Wavelet-based estimation methodology}

\subsection{Signal representation and wavelet domain model}
Let $x_1,\ldots,x_n$ denote equally spaced sampling locations of an acoustic signal, where $n=2^J$ for some integer $J$. The observed signal is modeled as
\begin{equation}
y_i=f(x_i)+\epsilon_i,\qquad i=1,\ldots,n,
\label{eq:regression_model}
\end{equation}
where $f$ is an unknown squared integrable function that represents the underlying bird vocalization and $\epsilon_i$ accounts for background noise and other sources of signal contamination. We assume that the noise variables are independent and identically distributed (iid) and satisfy
$$
    \epsilon_i\overset{\mathrm{iid}}{\sim}\mathcal{N}(0,\sigma^2),
$$
where $\sigma > 0$ is possibly unknown. In this case, we use a robust estimator of $\sigma$ based on the median absolute deviation of the empirical wavelet coefficients at the finest resolution level, as proposed by \cite{dj1994}. The objective is therefore to recover the relevant structure of $f$ from the noisy observations without imposing a specific parametric form on the acoustic signal.\\

\noindent Wavelets provide a convenient representation for this purpose because they describe a signal simultaneously according to scale and temporal location. In contrast to global representations, this property allows localized features of the vocalization, such as rapid changes in amplitude or oscillatory patterns, to be represented by a relatively small number of coefficients. Accordingly, we represent $f$ through a wavelet expansion of the form
\begin{equation}
f(x)=\sum_{j,k}\theta_{j,k}\psi_{j,k}(x),
\label{eq:wavelet_expansion}
\end{equation}
where
$$
    \psi_{j,k}(x)=2^{j/2}\psi(2^j x-k)
$$
is obtained from a wavelet function $\psi(x)$ by dilation and translation, and $\theta_{j,k}$ denotes the coefficient associated with scale $j = 0, \ldots, J-1$ and location $k = 0, \ldots, 2^{j} - 1$. For a suitable orthonormal wavelet system, these coefficients contain the information required to reconstruct the signal at different resolutions.\\

\noindent For the observed data, the discrete wavelet transform (DWT) provides a finite-dimensional representation of the sampled signal. Let $\boldsymbol{y}=(y_1,\ldots,y_n)^\top$ and $\boldsymbol{f}=(f(x_1),\ldots,f(x_n))^\top$. The observation model in vector form is
\begin{equation}
\boldsymbol{y}=\boldsymbol{f}+\boldsymbol{\epsilon},
\label{eq:vector_model}
\end{equation}
where $\boldsymbol{\epsilon}=(\epsilon_1,\ldots,\epsilon_n)^\top$.\\

\noindent Denote by $\boldsymbol{W}$ the orthogonal matrix associated with the chosen DWT. Applying the transform to the observations gives
\begin{equation}
\boldsymbol{d}=\boldsymbol{W}\boldsymbol{y}
=\boldsymbol{W}\boldsymbol{f}+\boldsymbol{W}\boldsymbol{\epsilon}
=\boldsymbol{\theta}+\boldsymbol{\eta},
\label{eq:wavelet_domain}
\end{equation}
where $\boldsymbol{d}$ contains the empirical wavelet coefficients, $\boldsymbol{\theta}=\boldsymbol{W}\boldsymbol{f}$ represents the corresponding coefficients of the underlying signal, and $\boldsymbol{\eta}=\boldsymbol{W}\boldsymbol{\epsilon}$ denotes the transformed noise.\\

\noindent Because $\boldsymbol{W}$ is orthogonal, the covariance structure of the independent Gaussian noise is preserved under the transformation. Consequently,

$$
    \boldsymbol{\eta}\sim\mathcal{N}(\boldsymbol{0},\sigma^2\boldsymbol{I}_n),
$$
where $\boldsymbol{I_n}$ is the $n \times n$ identity matrix and each transformed observation can be considered as a noisy measurement of its corresponding wavelet coefficient. Thus, for an individual coefficient, the estimation problem can be written in the simpler scalar form
\begin{equation}
d=\theta+\eta,
\qquad
\eta\sim\mathcal{N}(0,\sigma^2).
\label{eq:scalar_wavelet_model}
\end{equation}

\noindent The wavelet representation is particularly useful for denoising because coefficients associated with weak or noise-dominated components can be strongly attenuated, whereas coefficients carrying relevant signal information can be retained. In this work, this attenuation is performed through a Bayesian shrinkage rule based on the Epanechnikov prior described in the following subsection. The procedure is applied independently to the empirical wavelet coefficients, producing estimates $\widehat{\theta}_{j,k}$. Finally, the denoised acoustic signal is obtained by applying the inverse discrete wavelet transform (IDWT),
\begin{equation}
\widehat{\boldsymbol{f}}
=\boldsymbol{W}^{\top}\widehat{\boldsymbol{\theta}}.
\label{eq:idwt}
\end{equation}
More details about wavelet representation of signals and the DWT, see \cite{vidakovic1999}.

\subsection{Bayesian wavelet shrinkage rule}
\noindent Wavelet shrinkage estimation is typically performed by reducing the magnitude of the empirical wavelet coefficients to obtain estimates of the true coefficients. A variety of shrinkage rules have been proposed in the literature, most of which are based on thresholding procedures. These methods set empirical coefficients to zero whenever their absolute values fall below a prescribed threshold, while retaining or shrinking larger coefficients. Seminal contributions to this area include the works of \citep{dj1994} and \citep{dj1995}.\\

\noindent Under a Bayesian perspective, it is possible to incorporate prior information about the parameters by a prior distribution. A common prior $\pi(\cdot)$ for a single wavelet coefficient $\theta$ is a mixture of a point mass function at zero $\delta_0(\cdot)$ and a symmetric around zero and unimodal probability density function $g(\cdot;\beta)$, i.e 
\begin{equation}
    \label{prior2}
    \pi(\theta;\alpha,\beta)=\alpha \delta_0(\theta)+(1-\alpha)g(\theta;\beta),
\end{equation}
where $\alpha \in (0,1)$ and $\beta$ are hyperparameters. In this work, we consider the prior distribution based on the Epanechnikov kernel function proposed by \citep{fidel} given by \eqref{prior2} with $g(\cdot)$ given by
\begin{align}
\label{prior1}
     g(\theta;\beta)=\frac{3}{4\beta^3}(\beta^2-\theta^2)\mathbb{I}_{(-\beta,\beta)}(\theta),
\end{align}

\noindent where $\beta > 0$ and $\mathbb{I}_{\{A\}}(\cdot)$ is the indicator function on the set $A$. Further, it is assumed an exponential distribution as prior distribution for $\sigma^2$,
\begin{equation}\label{prior3}
    \pi (\sigma^2;\lambda) = \lambda e^{-\lambda\sigma^2} \mathbb{I}_{(0,\infty)}(\sigma^2), 
\end{equation}
$\lambda>0$. Thus, under the models \eqref{wavelet_model}, \eqref{prior2}, \eqref{prior1} and \eqref{prior3} and the squared loss function $L(\theta,\delta)=(\delta-\theta)^2$, the associated shrinkage rule is the posterior mean of $\theta$, i.e,
 \small
\begin{align}
\label{rule} 
\delta(d)=& \mathbb{E}(\theta \mid d) \\
=& \frac{(1-\alpha) \frac{3\sqrt{ 2 \lambda } }{ 8\beta^3 }\left[ \frac{2\lambda\beta^2 +3\sqrt{2\lambda}\beta+3}{2\lambda^2}\left(e^{-\sqrt{2\lambda}(\beta-d)}- e^{-\sqrt{2\lambda}(\beta+d)}\right)+ \frac{(\lambda\beta^2 -3)\sqrt{2\lambda}d -\lambda\sqrt{2\lambda}d^3}{\lambda^2}\right]}{\alpha \mathcal{ED}\left(0,\frac{1}{\sqrt{2\lambda}}\right)+(1-\alpha) \frac{3\sqrt{2\lambda}}{8\beta^3}\left[ \frac{\beta}{\lambda}\left( e^{-\sqrt{2\lambda}(\beta+d)} + e^{-\sqrt{2\lambda}(\beta-d)} \right)+ \frac{2}{\sqrt{2\lambda}}\left( \beta^2 - d^2 -\frac{1}{\lambda}\right)\right]},\nonumber
\end{align}

\begin{figure}[H]
\centering
\includegraphics[width=0.7\textwidth]{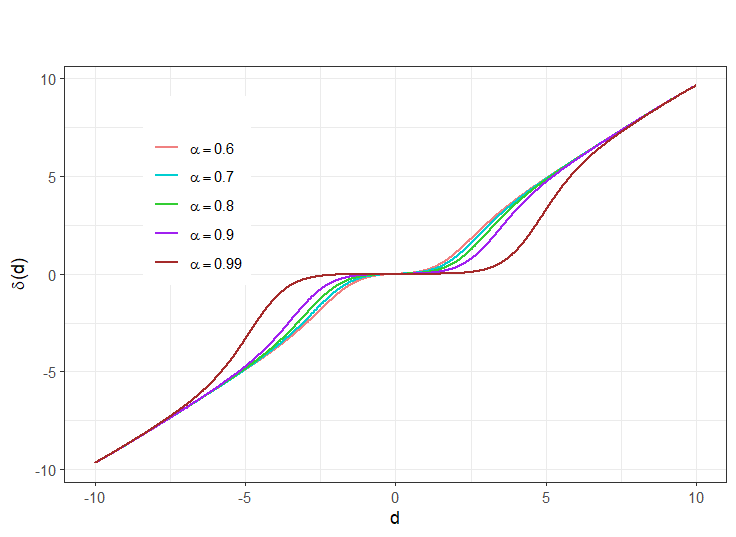}
\caption{Wavelet shrinkage rule \eqref{rule} for $\beta=12$, $\lambda=1.3$ and several values of $\alpha$.}
\label{gRuler}
\end{figure}

\small

\noindent where \( \mathcal{ED}\left(0, \frac{1}{\sqrt{2\lambda}}\right) \) is the probability density function of the double exponential distribution with mean equals to zero and scale parameter equals to \( \frac{1}{\sqrt{2\lambda}} \). Figure \ref{gRuler} shows the shrinkage rule \eqref{rule} for $\beta = 12$, $\lambda = 1.3$ and several values of $\alpha$. In fact, the hyperparameter $\alpha$ has an impact on the severity of the shrinkage rule, which reduces the magnitude of the empirical coefficient. See \citep{vidakovic1999} and \citep{nason2008} for a general overview on Bayesian wavelet shrinkage.


\noindent The choice of the Epanechnikov rule in this study was motivated by both its theoretical properties and its empirical performance in function estimation problems under noisy conditions. This rule is based on a simple prior distribution that has been widely used in the statistical literature, facilitating interpretation and application across different settings. Furthermore, the adopted formulation leads to an explicit decision rule, avoiding complex iterative procedures and resulting in high computational efficiency a particularly important feature when processing large collections of bioacoustic recordings. Another relevant aspect is that previous studies have shown that the Epanechnikov rule performs competitively and often outperforms alternative approaches in low signal-to-noise ratio scenarios, a condition frequently encountered in field recordings of bird songs. Across several simulation studies and practical applications, this method has surpassed both classical estimation rules and alternative Bayesian procedures available in the literature, including methods specifically designed for low signal-to-noise ratio settings, such as the Gamma Minimax rule based on three-point prior distributions. Therefore, its use in the present study is well justified for recovering the underlying bioacoustic signal, combining estimation accuracy, robustness to noise, and computational efficiency.

\subsection{Hyperparameters elicitation}

The specification of the hyperparameters associated with the Epanechnikov rule plays a crucial role in the estimation procedure, since these parameters determine the amount of shrinkage applied to the wavelet coefficients and directly affect the ability of the method to separate the underlying bioacoustic signal from background noise. In this study, the adopted specifications were based on previous literature and on evidence of good performance in low signal-to-noise ratio settings, a common characteristic of bird-song recordings. The hyperparameter $\alpha$ was defined as a resolution-level dependent quantity according to
\begin{equation}\label{alpha}
    \alpha=\alpha(j)=1-\frac{1}{(j-J_0+l)^\gamma},
\end{equation}

\noindent where $J_0$ is the primary resolution level and $J_0 \leq j \leq J-1$. Throughout the study, the shrinkage rule was applied to the entire vector of empirical coefficients, i.e., $J_0 = 0$. Following the recommendation of \cite{vimalajeewa2023}, the values $l=1$ and $\gamma=2$ were adopted. This specification yields increasing values of $\alpha(j)$ as the resolution level increases, resulting in stronger shrinkage of coefficients associated with finer scales, where a substantial portion of the noise is typically concentrated. The hyperparameter $\beta$ was selected adaptively at each resolution level according to
\begin{equation}
\label{beta}
    \beta=\beta(j)=\underset{k}{\mathrm{max}} \left\lvert d_{j,k}\right\rvert,
\end{equation}

\noindent where the maximization is performed over all wavelet coefficients belonging to level $j$. This choice allows the support of the prior distribution, defined on the interval $(-\beta,\beta)$, to adjust automatically to the observed magnitude of the coefficients at each scale, providing greater flexibility in the estimation process. Finally, the hyperparameter $\lambda$ was determined from the variability of the wavelet coefficients at the finest resolution level through
\begin{equation}
\label{lambda}
   \lambda=\lambda(s)=\frac{1}{s^2}+\frac{c}{\tau} \exp\left(-\frac{1}{\tau} s\right),
\end{equation}

\noindent where $s$ denotes the sample standard deviation of the wavelet coefficients at the highest resolution level. In the absence of prior information, the values $c=1$ and $\tau=2$ were adopted. This specification enables $\lambda$ to adapt to the noise level present in the recording, providing stronger smoothing when the observed variability is high while preserving relevant acoustic structures when the signal-to-noise ratio is more favorable. See \cite{fidel} for more details.


\section{Materials and methods}
\subsection{\textit{Euphonia violacea}}
The violaceous euphonia (\textit{Euphonia violacea}) is a resident passerine widely distributed across northern and eastern South America, ranging from Venezuela and Trinidad to Paraguay, northeastern Argentina, and southeastern Brazil. It occupies a variety of habitats, including humid forests, forest edges, parks, gardens, cocoa plantations, and citrus orchards. Sexual dimorphism is pronounced in this species: males exhibit glossy violet-blue to blackish upperparts and bright golden-yellow underparts, whereas females and juveniles display predominantly olive-green and yellowish-olive plumage. Its diet consists mainly of fruits, although nectar and insects may also be consumed when seasonally available. Individuals are typically observed alone, in pairs, or in mixed-species foraging flocks \citep{martinez2020}.

\subsection{\textit{Leiothrix lutea}}
Over recent decades, the red-billed leiothrix (\textit{Leiothrix lutea}), a species native to Asia, has successfully established populations in several European countries following accidental escapes and intentional releases. Although studies conducted in other invaded regions have suggested that this species may negatively affect local ecosystems, its ecological impacts in Europe remain insufficiently understood. Evidence indicates that \textit{Leiothrix lutea} may compete with native bird species and potentially alter community structure through its competitive dominance. According to \cite{leiothrix}, records of the species were documented across 37 regions in 10 European countries, with self-sustaining populations confirmed in France, Italy, Spain, and Portugal. Furthermore, the species' European distribution expanded substantially, nearly doubling within a period of less than two decades. Species distribution modeling revealed that the likelihood of occurrence was positively associated with human population density, precipitation seasonality, precipitation during the driest quarter, minimum temperature of the coldest month, and the spatial distribution of previous occurrence records.

\subsubsection{Risk analysis}
Public and academic recognition of the problems associated with biological invasions has grown exponentially over the past decade. The reasons for this growth are three-fold. First, the negative effects of some non-native species have grown too large to ignore. Thus, increasing numbers of scientists are studying and managing non-native species in an effort to minimize the effects of biological invaders on native species and human economies. Second, the number of species being moved out of their native ranges and into novel locations is itself growing. Therefore, not only are the problems caused by non-native species becoming blatantly obvious, but also the overall number of problems appears to be growing. Third, with so many invasive species, it is very hard to do
ecological field research without encountering invaders and eventually including them in investigations even if those investigations are for basic research. Invaders offer some new insights, and it is very difficult for curious scientists to pass up the opportunity to explore these new avenues \citep{invasion}. With \href{https://drive.google.com/file/d/1wemsLrvr7uYfKNZjJa5QwjL2qEesVjYC/view?usp=sharing}{form} we made the Risk analysis of \textit{Leiothrix lutea}.

\begin{table}[H]
\centering
\caption{Risk assessment summary for the invasive bird species \textit{Leiothrix lutea}. The table presents the final risk score obtained from the assessment protocol, including the number of responses in each evaluation section: (A) Biological and ecological characteristics, (B) Biogeographical aspects, (C) Social and economic aspects, and (D) Risk-enhancing characteristics. The species was classified as presenting a very high invasion risk, resulting in a recommendation for rejection.}
\label{tab:risk}
\begin{tabular}{lc}
\hline
Taxon / Species                   & \multicolumn{1}{l}{\textit{Leiothrix lutea}} \\\hline
Final score                       & 81.5                     \\
Section A responses               & 16                   \\
Section B responses               & 5  \\
Section C responses               & 6  \\
Section D responses               & 12  \\
Total responses                   & 39  \\
Minimum criteria met              & \textcolor{green}{Valid RA}                          \\
Potential risk                    & Very high                          \\
Recommendation                    & \textcolor{red}{Reject}    \\\hline                        
\end{tabular}
\end{table}

 \begin{figure}[H]
\centering
\includegraphics[width=0.6\textwidth]{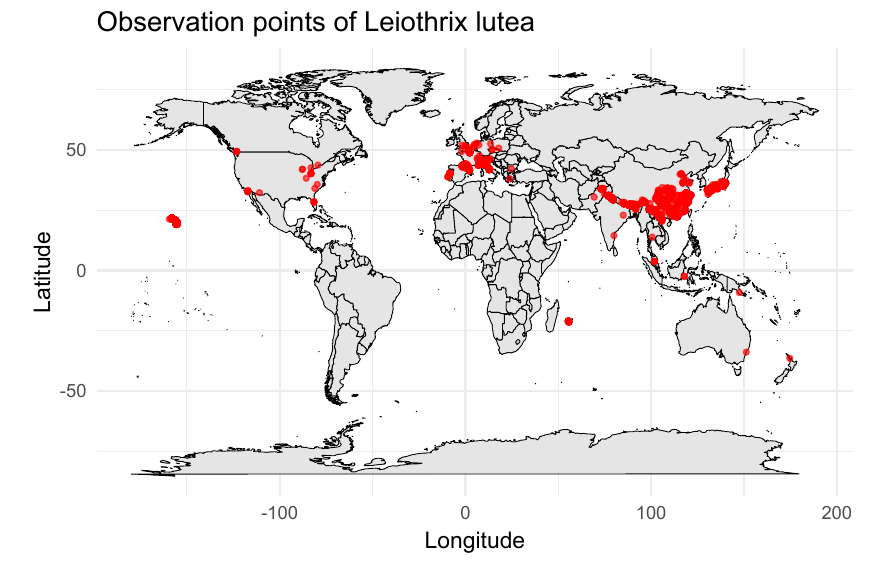}
\caption{Geographic distribution of \textit{Leiothrix lutea} observations in the world, based on \href{https://www.inaturalist.org/photos/62222232}{iNaturalist} data (n = 5000 records).}\label{mapa1}
\end{figure} 
\noindent The risk assessment conducted for \textit{Leiothrix lutea} resulted in a final score of 81.5 (see Table \ref{tab:risk}), classifying the species as presenting a very high invasion risk. The species met the minimum criteria required for a valid assessment, with responses distributed across all evaluation sections. High scores in the biological and ecological criteria indicate a strong capacity for establishment, survival, and spread in non-native environments. In addition, the species showed relevant biogeographical and socio-economic risk factors, as well as several characteristics associated with invasion success (Figure \ref{mapa1}). Based on the assessment protocol, the species received a recommendation for rejection, indicating that its introduction or management should be restricted due to its high invasive potential and possible negative impacts on native biodiversity and ecosystems.

\subsection{\textit{Passer domesticus}}
The house sparrow (\textit{Passer domesticus}) is a bird of the sparrow family Passeridae, found in most parts of the world. One of about 25 species in the genus Passer, the house sparrow is native to most of Europe, the Mediterranean Basin, and a large part of Asia. Its intentional or accidental introductions to many regions, including parts of Australasia, Africa, and the Americas, make it the most widely distributed wild bird. The house sparrow is host to a huge number of parasites and diseases, and the effect of most is unknown. The commonly recorded bacterial pathogens of the house sparrow are often those common in humans, and include Salmonella and Escherichia coli. Salmonella is common in the house sparrow, and a comprehensive study of house sparrow disease found it in 13\% of sparrows tested \citep{passer}.

\subsubsection{Risk analysis}
\noindent Table \ref{tab:risk2} summarizes the risk assessment results for the invasive bird species \textit{Passer domesticus}. The species obtained a final score of 91.5, classifying it as a species with \textit{very high} invasion risk according to the adopted assessment protocol. This elevated score indicates a strong combination of biological, ecological, biogeographical, and socio-economic characteristics associated with invasion success and potential environmental impact. Importantly, the assessment satisfied the minimum criteria required for a valid risk analysis (“Valid RA”), demonstrating that the available evidence was sufficient to support a reliable classification. Based on the overall score and the associated risk category, the final recommendation was “Reject”, indicating that the introduction, maintenance, or further spread of the species should not be encouraged due to its considerable invasive potential and associated ecological risks.
\begin{table}[H]
\centering
\caption{Risk assessment summary for the invasive bird species \textit{Passer domesticus}. The table presents the final risk score obtained from the assessment protocol, including the number of responses in each evaluation section: (A) Biological and ecological characteristics, (B) Biogeographical aspects, (C) Social and economic aspects, and (D) Risk-enhancing characteristics. }
\label{tab:risk2}
\begin{tabular}{lc}
\hline
Taxon / Species                   & \multicolumn{1}{l}{\textit{Passer domesticus}} \\\hline
Final score                       & 91.5                     \\
Section A responses               & 16                   \\
Section B responses               & 5  \\
Section C responses               & 6  \\
Section D responses               & 12  \\
Total responses                   & 39  \\
Minimum criteria met              & \textcolor{green}{Valid RA}                          \\
Potential risk                    & Very high                          \\
Recommendation                    & \textcolor{red}{Reject}    \\\hline                        
\end{tabular}
\end{table}
 \begin{figure}[H]
\centering
\includegraphics[width=0.6\textwidth]{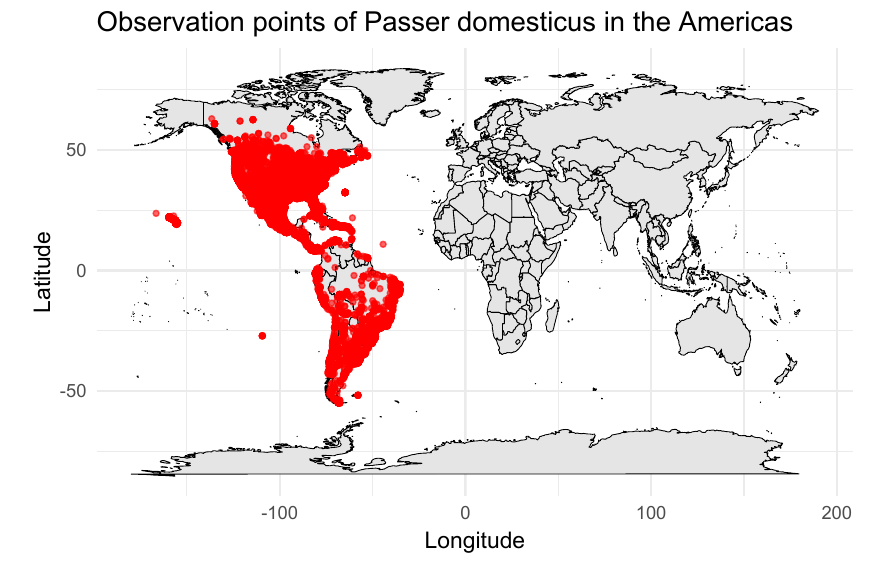}
\caption{Geographic distribution of \textit{Passer domesticus} observations in the Americas, based on \href{https://www.inaturalist.org/photos/62222232}{iNaturalist} data (n = 5000 records, years 2000 until 2025).}\label{mapa2}
\end{figure} 

\noindent \cite{Mustafa} the iNaturalist Sounds Dataset (iNatSounds) is a collection of $230,000$ audio files capturing sounds from over $5,500$ species, contributed by more than 27,000 recordists worldwide. The dataset encompasses sounds from birds, mammals, insects, reptiles, and amphibians, with audio and species labels derived from observations submitted to iNaturalist, a global citizen science platform. \\

\begin{figure}[H]
\centering
\includegraphics[width=0.75\textwidth]{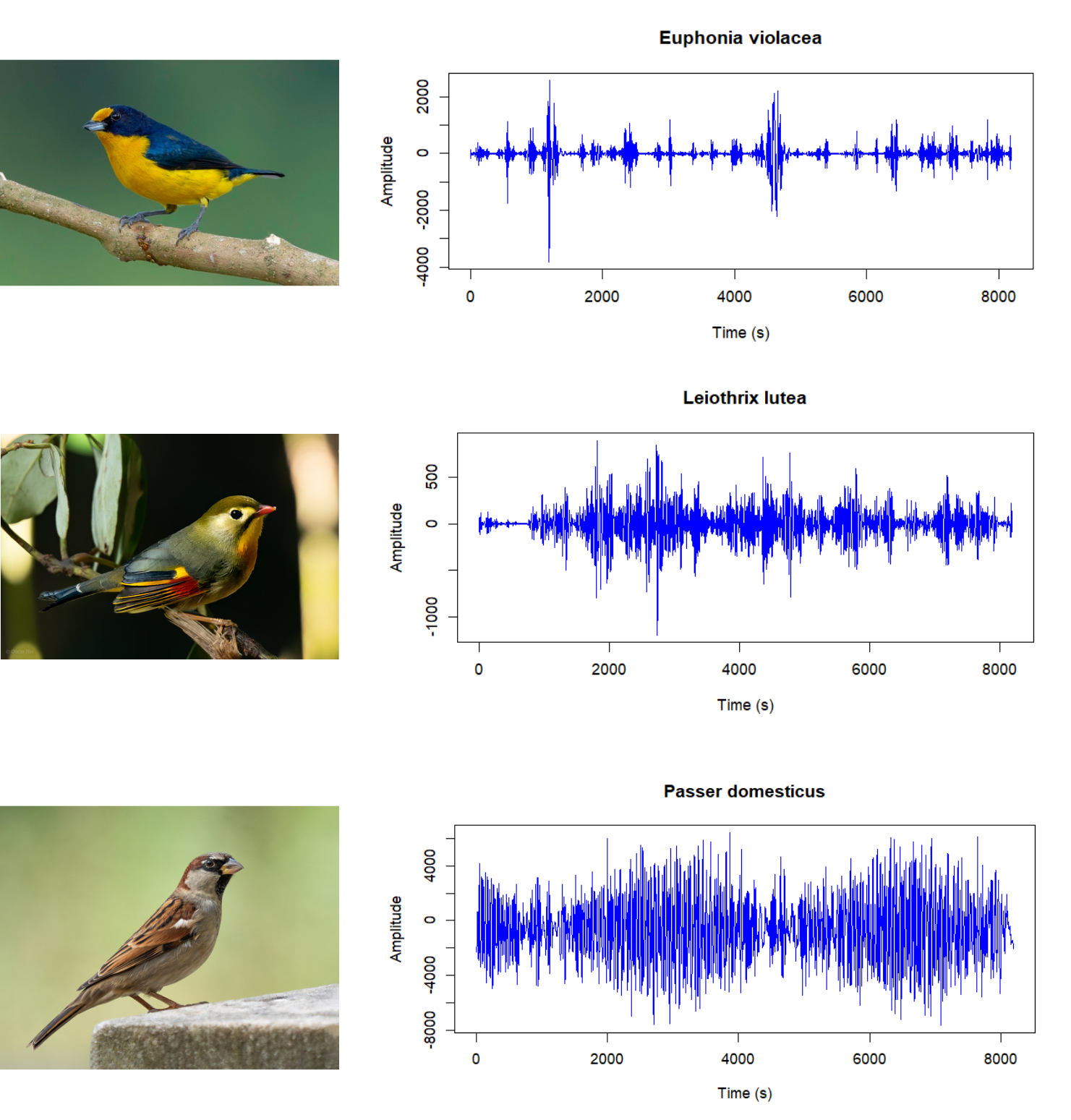}
\caption{Animals with frequency in R by \href{https://www.inaturalist.org/photos/28179600}{iNaturalist-\textit{Euphonia violacea}}, \href{https://www.inaturalist.org/photos/62222232}{iNaturalist-\textit{Leiothrix lutea}} and \href{https://www.inaturalist.org/photos/452209454}{iNaturalist-\textit{Passer domesticus}}}\label{brids}
\end{figure}

\noindent Acoustic data were collected from the iNaturalist platform using the R package \texttt{rinat}. Observations were retrieved for three major species groups: birds \textit{Euphonia violacea, Leiothrix lutea}, and \textit{Passer domesticus}, based on the query term “song” (See Figure \ref{brids}). Bird recordings were restricted to the year 2025. Records without valid audio files were excluded. All datasets were combined into a single database, and the variable iconic$\_$taxon$\_$name was used to define the classification groups.\\

\noindent Audio files were downloaded directly from their URLs using the \texttt{httr} package. Given the heterogeneity of file formats, a two-stage procedure was adopted to ensure successful decoding. First, audio files were converted to WAV format using \texttt{av}; when this step failed, a fallback approach based on MP3 decoding via \texttt{tuneR} was employed. All signals were resampled to 16 kHz to ensure consistency across recordings. Silence segments were removed using functions from the \texttt{seewave} package, thereby reducing non-informative portions of the signals. Each processed signal was then truncated to a length equal to the largest power of two not exceeding its original size, ensuring compatibility with wavelet-based transformations.
\begin{tcolorbox}[colback=pink!5!white,colframe=pink!75!black,fonttitle=\bfseries,title=Interquartile Range (IQR)]
 Is a measure of statistical dispersion that describes the spread of the middle 50 \% of a dataset, calculated by subtracting the first quartile (Q1) from the third quartile (Q3). It represents the range between the 25th and 75th percentiles and is a robust measure of variability often used to identify outliers.
\end{tcolorbox}

\begin{tcolorbox}[colback=pink!5!white,colframe=pink!75!black,fonttitle=\bfseries,title=Mel-frequency cepstral coefficients (MFCC)]
Are widely used in bioacoustics to represent bird vocalizations by extracting key spectral features that mimic how sound is perceived. They are crucial for automated bird species recognition, species classification and monitoring \citep{MFCC}.
\end{tcolorbox}

\begin{tcolorbox}[colback=pink!5!white,colframe=pink!75!black,fonttitle=\bfseries,title=Root Mean Square (RMS)]
Is a statistical measure of the magnitude of a varying quantity, calculated as the square root of the mean of the squares of a set of values.
\end{tcolorbox}

\begin{tcolorbox}[colback=pink!5!white,colframe=pink!75!black,fonttitle=\bfseries,title=Spectral centroid (SC)]
 In birds, it is an acoustic measure that indicates the "center of gravity" of the power spectrum of a vocalization, essentially determining where most of the sound energy is concentrated (whether it is higher or lower pitch).
\end{tcolorbox}

\begin{tcolorbox}[colback=pink!5!white,colframe=pink!75!black,fonttitle=\bfseries,title=Spectral Entropy (Entropy-SP)]
In birdsong analysis is a measure of the complexity, disorder, or "noisiness" of a sound based on its power spectrum, often calculated using the Shannon entropy formula. It quantifies how energy is distributed across frequencies in a signal \citep{SP}.
\end{tcolorbox}

\begin{tcolorbox}[colback=pink!5!white,colframe=pink!75!black,fonttitle=\bfseries,title=Spectral centroid (Centroid)]
 In birdsong analysis represents the \texttt{center of mass} or weighted average frequency of a vocalization, acting as a proxy for perceived sound \texttt{brightness} or \texttt{warmth}. It indicates where most of the sound energy is concentrated in the frequency spectrum, with higher centroid values corresponding to brighter, higher-pitched sounds \citep{centroid}.
\end{tcolorbox}

\begin{tcolorbox}[colback=pink!5!white,colframe=pink!75!black,fonttitle=\bfseries,title=Zero Crossing Rate (ZCR)]
In bird sounds measures the frequency at which a bird’s audio signal passes through the zero-amplitude axis, providing a computational metric to analyze the spectral content of a call. It is widely used in ornithology and bioacoustics as a time-domain feature for analyzing the rapid, high-frequency nature of bird songs \citep{ZCR}.
\end{tcolorbox}

\subsection{Statistical analysis}

\noindent To reduce noise while preserving relevant acoustic structures, a Bayesian Wavelet denoising approach was applied through a custom implementation. This method operates by Shrinking Wavelet coefficients under a prior structure controlled by hyperparameters, allowing adaptive smoothing of the signal while maintaining important temporal features. The resulting denoised signals formed the basis for subsequent feature extraction.\\

\noindent Feature extraction was performed using discrete Wavelet decomposition implemented in the Wavelets. A Daubechies wavelet filter (DaubExPhase, filter number 10) was applied to each signal. For each decomposition level, energy-based descriptors were computed, including both the total energy and the mean energy of the detail coefficients. Additionally, the standard deviation of each denoised signal was included as a global measure of variability. These features were combined to form a structured dataset representing the multiscale characteristics of each acoustic signal.\\

\noindent The resulting dataset was partitioned into training and testing subsets using stratified sampling to preserve class proportions, as implemented in the caret package. Specifically, 80\% of the data were allocated to model training and 20\% to testing through a Monte Carlo cross-validation scheme with 200 independent random resampling iterations. Within each iteration, model training and performance assessment were conducted independently, allowing the evaluation to capture variability associated with different train–test splits and to provide robust estimates of predictive performance.\\

\noindent Because the classification problem involved two bird species and class imbalance could affect model fitting, random down-sampling was applied to the training data at each iteration. The majority class was reduced to match the frequency of the minority class, ensuring balanced class representation during model training while leaving the validation data unchanged to preserve the original class distribution for performance assessment.\\

\noindent Several classification models were then trained and compared: Random Forest (RF), Relevance Vector Machine (RVM), k-Nearest Neighbors (KNN), Support Vector Machine with radial basis function kernel (SVM-RBF), and multinomial logistic regression. A Random Forest classifier was fitted using the \textit{randomForest} package with hyperparameter tuning performed via 10-fold cross-validation. For KNN and SVM-RBF, predictor variables were centered and scaled before model fitting, and tuning parameters were selected through 10-fold cross-validation. A Multinomial Logistic regression model was also fitted, with predictors standardized prior to estimation and model selection conducted using 5-fold cross-validation. For the RVM approach, a radial basis kernel was adopted, and predicted scores were transformed into posterior probabilities using the logistic function, with class assignments obtained using a threshold of 0.50. Finally, a Support Vector Machine with radial kernel was trained using the \textit{e1071}, with hyperparameters selected via cross-validation and predictors standardized \citep{hastie2009elements}.\\

\noindent Model performance was evaluated on the independent validation subset at each iteration. Classification results were summarized using confusion matrices, from which several performance metrics were derived, including overall accuracy, sensitivity, specificity, balanced accuracy, and Cohen’s kappa coefficient. In addition, McNemar’s test was computed to assess potential asymmetries in classification errors. The final performance of each algorithm was reported as the average of the corresponding metric across the 200 resampling iterations, providing stable estimates of predictive accuracy and classification reliability.\\

\noindent A nested validation strategy was therefore adopted, in which hyperparameter tuning was performed exclusively within the training data using 10-fold cross-validation, whereas predictive performance was assessed on an external validation subset not used during model fitting. This procedure reduces the risk of optimistic bias and provides a more realistic estimate of model generalization ability when applied to unseen observations. All analyses were conducted within the \textbf{R} statistical computing environment. All codes are in the repository of \href{https://github.com/lldb14/Bird-Wavelet}{GitHub}.

\begin{figure}[H]
\centering

\begin{subfigure}{0.4\textwidth}
    \centering
\includegraphics[width=7.3cm,height=8.8cm,keepaspectratio]{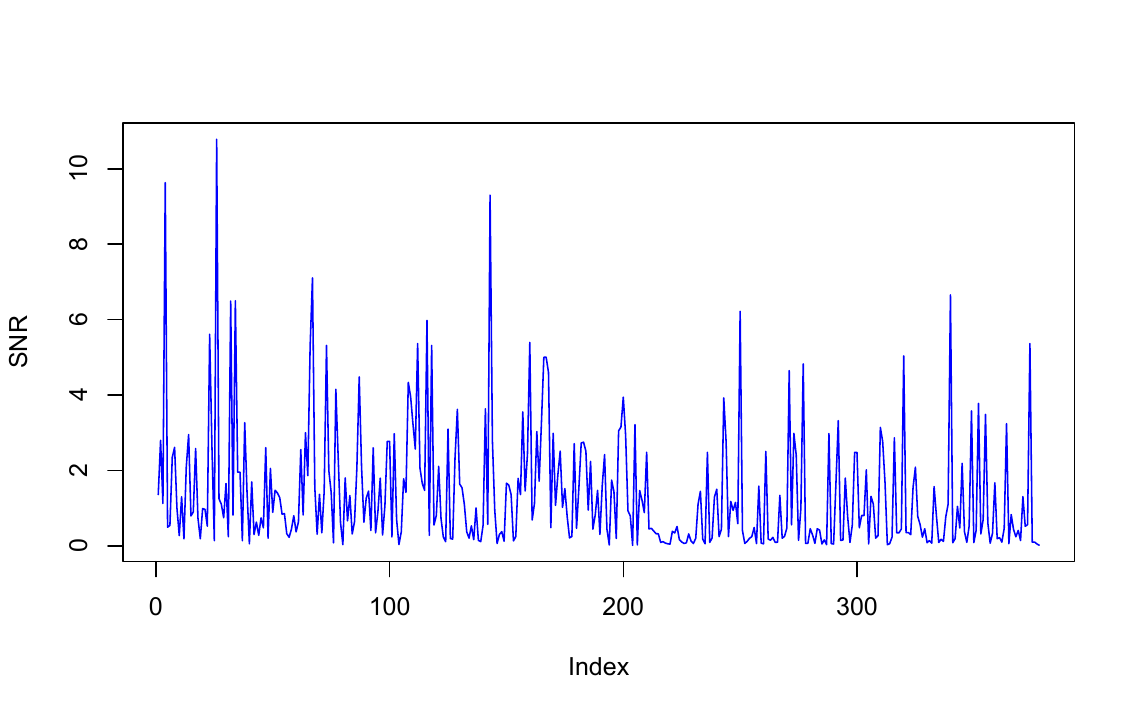}
    \caption{Data 1 - \textit{Passer domesticus} and \textit{Leiothrix lutea}}
    \label{m5}
\end{subfigure}\hspace{0.1\textwidth} 
\begin{subfigure}{0.44\textwidth}
    \centering
\includegraphics[width=7.3cm,height=8.8cm,keepaspectratio]{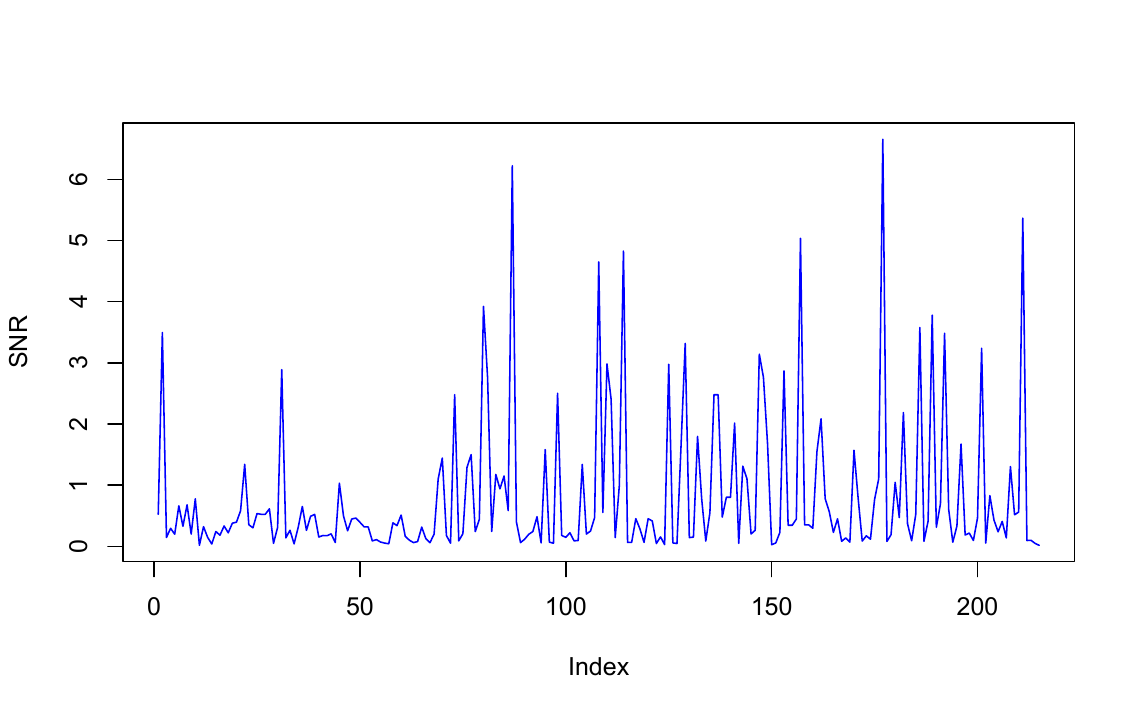}
    \caption{Data 2 - \textit{Euphonia violacea} and \textit{Leiothrix lutea}}
    \label{m10}
\end{subfigure}
\caption{Typical SNR Distribution in Birdsong Recordings. The signal-to-noise ratio (SNR) in bioacoustic signals}
\label{distris}
\end{figure}

\noindent Figure \ref{distris} illustrates the temporal distribution of the signal-to-noise ratio (SNR) across birdsong recordings for the two datasets. In both datasets, the SNR values exhibit substantial variability, characterized by intermittent peaks interspersed with periods of relatively low signal intensity. This pattern reflects the heterogeneous acoustic conditions commonly observed in bioacoustic recordings, where vocalizations are influenced by environmental noise, recording distance, and background interference.\\

\noindent For Data 1 (\textit{Passer domesticus} and \textit{Leiothrix lutea}), the SNR distribution shows a broader range of fluctuations, including several pronounced peaks exceeding 10 dB. These results suggest the presence of recordings with highly distinguishable vocal signals, although many segments still exhibit low SNR values, indicating challenging acoustic environments. In contrast, Data 2 (\textit{Euphonia violacea} and \textit{Leiothrix lutea}) presents a comparatively narrower SNR range, with most values concentrated at lower levels and fewer extreme peaks. Nevertheless, sporadic increases in SNR are also evident, demonstrating occasional segments with clearer vocal activity.

\section{Results and discussion}
\begin{figure}[H]
\centering
\begin{subfigure}{0.4\textwidth}
    \centering
\includegraphics[width=5.2cm,height=6.2cm,keepaspectratio]{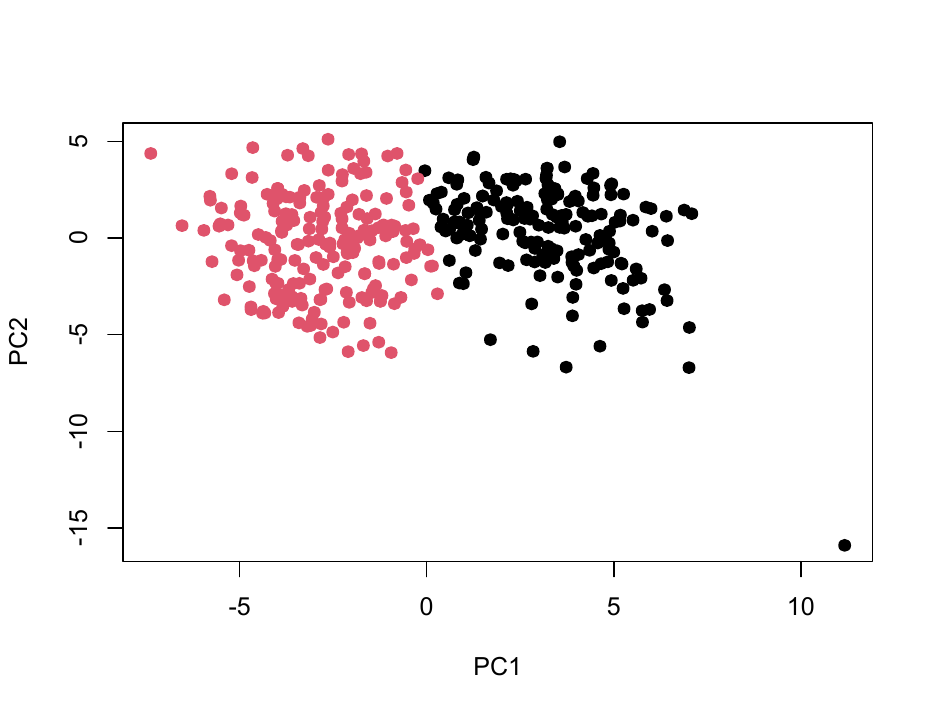}
    \caption{m=4 Data 1 - \textit{Passer domesticus} and \textit{Leiothrix lutea}}
    \label{m5pl}
\end{subfigure}\hspace{0.1\textwidth} 
\begin{subfigure}{0.4\textwidth}
    \centering
\includegraphics[width=5.2cm,height=6.2cm,keepaspectratio]{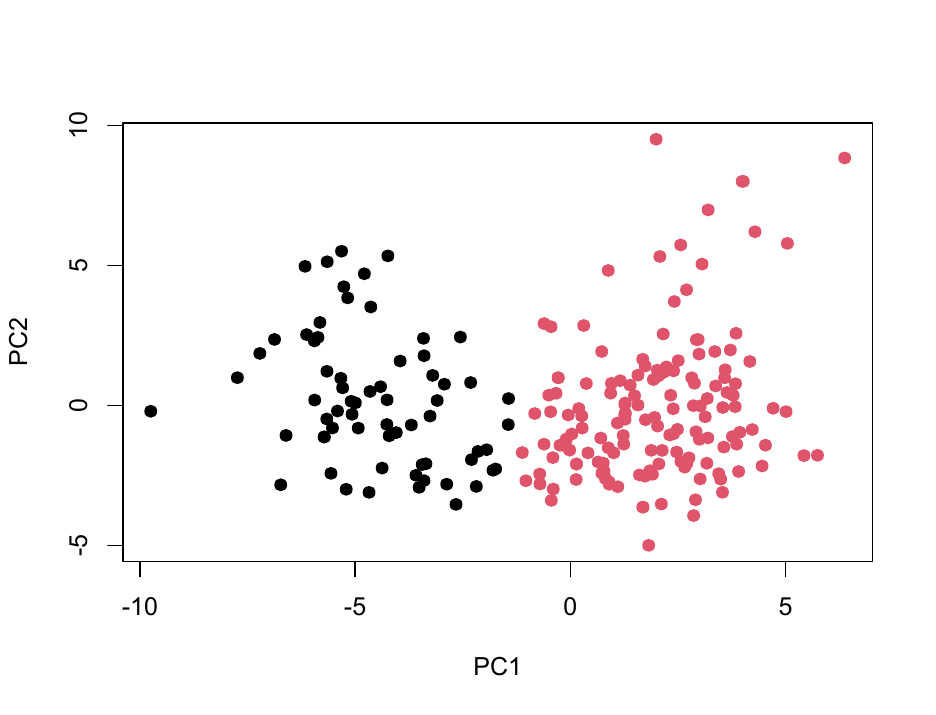}
    \caption{m=4 Data 2 - \textit{Euphonia violacea} and \textit{Leiothrix lutea}}
    \label{m4el}
\end{subfigure}
\begin{subfigure}{0.4\textwidth}
    \centering
\includegraphics[width=5.2cm,height=6.5cm,keepaspectratio]{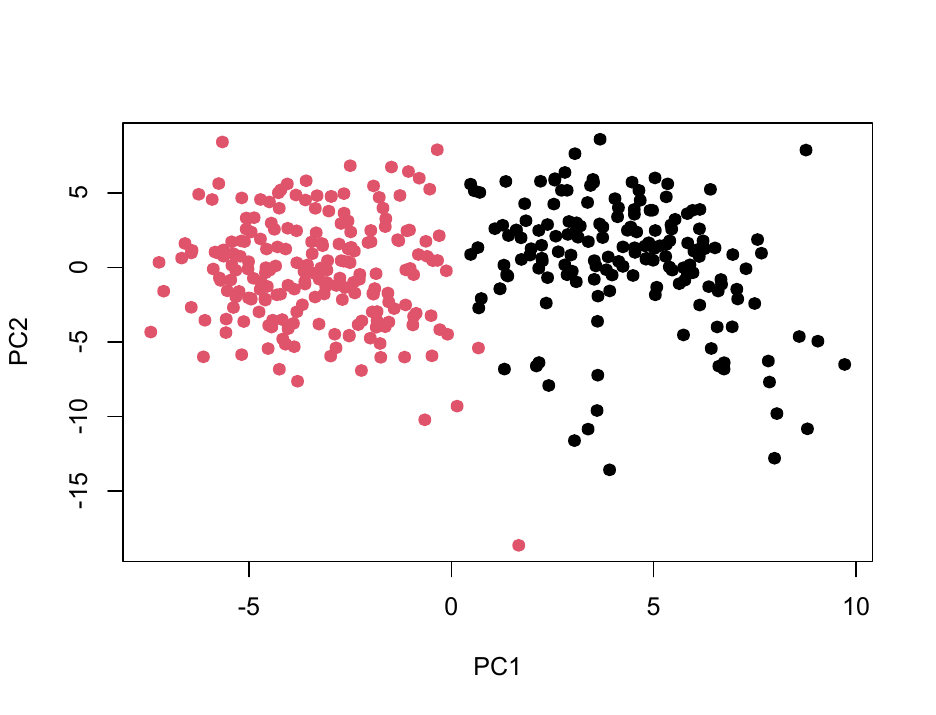}
    \caption{m=10 Data 1 - \textit{Passer domesticus} and \textit{Leiothrix lutea}}
    \label{m10pl}
\end{subfigure}\hspace{0.1\textwidth} 
\begin{subfigure}{0.4\textwidth}
    \centering
\includegraphics[width=5.2cm,height=6.5cm,keepaspectratio]{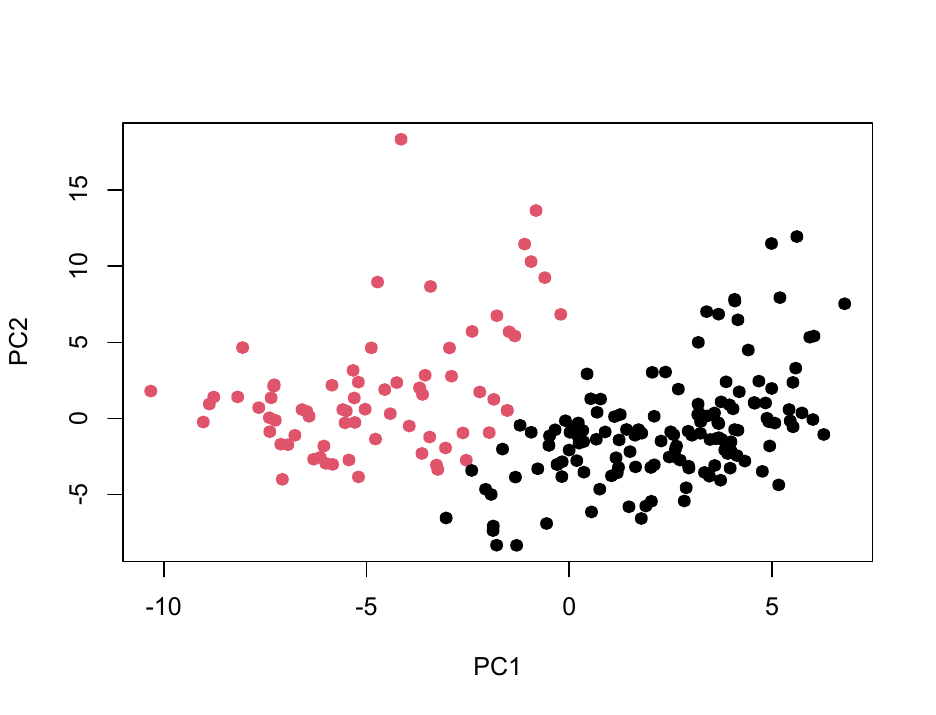}
    \caption{m=10 Data 2 - \textit{Euphonia violacea} and \textit{Leiothrix lutea}}
    \label{m10el}
\end{subfigure}
\caption{Principal Component Analysis (PCA) projection of the acoustic feature space for species discrimination. Plots (\ref{m5pl}) and (\ref{m10pl}) correspond to Data 1 (Passer domesticus and Leiothrix lutea) for feature dimensionalities m=4 and m=10, respectively. Plots (\ref{m4el}) and (\ref{m10el}) correspond to Data 2 (Euphonia violacea and Leiothrix lutea) for m=4 and m=10.}
\label{pca}
\end{figure}

The Figure \ref{pca} shows that the species exhibit distinct acoustic patterns that can be effectively captured by the first principal components. The separation is particularly pronounced when comparing Euphonia violacea and Leiothrix lutea, while the comparison between Passer domesticus and Leiothrix lutea shows a slight overlap, indicating a greater relative similarity between their acoustic characteristics. The stability of the observed patterns between m=4 and m=10 supports the robustness of the variable selection procedure and its ability to preserve biologically relevant information for species classification \cite{Seewave}.\\

\noindent In Figures (\ref{m5pl}) and (\ref{m10pl}), corresponding to m=4 and m=10, respectively, a clear separation between the two species is observed, primarily along the first principal component (PC1). \textit{Passer domesticus} individuals are concentrated at negative PC1 values, while \textit{Leiothrix lutea} individuals are clustered at positive values. Although there is a slight overlap in the central region of the graph, the overall structure of the groups remains well-defined for both m values. Increasing m from 4 to 10 does not substantially alter the separation between species, suggesting that the main discriminating information is already captured with a reduced number of features. However, at m=10, greater dispersion is observed within \textit{Leiothrix lutea}, especially in the PC2 direction, which could indicate greater intraspecific variability or the inclusion of additional, less discriminating variables.\\

\noindent Figures (\ref{m4el}) and (\ref{m10el}) show a more pronounced separation between groups than that observed in Data 1. At both m values, the two species form clearly differentiated clusters with minimal overlap. Discrimination occurs mainly along PC1, while PC2 captures internal variability within each species. For m=4 (Figure \ref{m4el}), the groups appear compact and relatively homogeneous. As m increases to 10 (Figure \ref{m10el}), the separation between centroids is maintained, although an increase in internal dispersion and the appearance of some extreme values are observed, particularly in the group corresponding to \textit{Euphonia violacea}. Despite this, the structure of the groups remains clearly distinguishable.\\

\begin{figure}[H]
\centering

\begin{subfigure}{0.85\textwidth}
    \centering
\includegraphics[width=10.5cm,height=10.8cm,keepaspectratio]{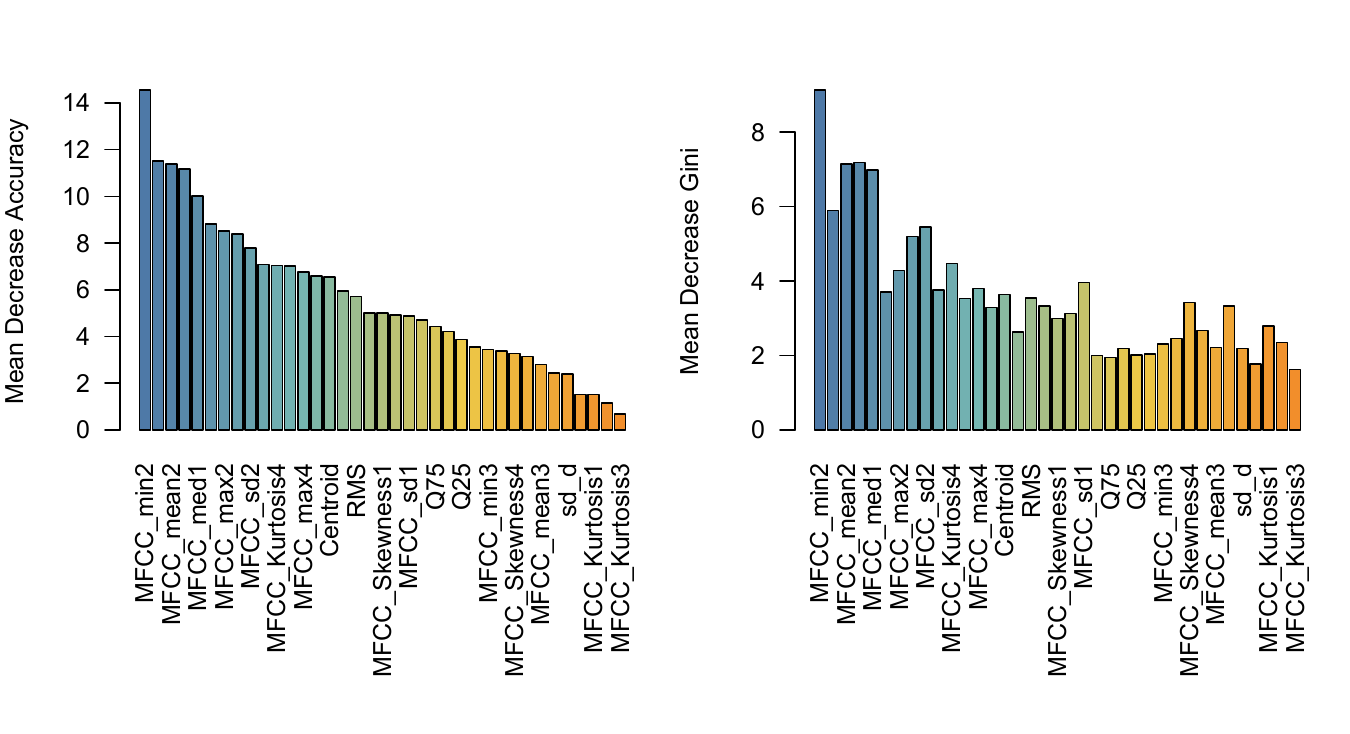}
    \caption{m=4 Data 1 - \textit{Passer domesticus} and \textit{Leiothrix lutea}}
    \label{m4}
\end{subfigure}\hspace{0.2\textwidth} 
\begin{subfigure}{0.85\textwidth}
    \centering
\includegraphics[width=10.5cm,height=10.8cm,keepaspectratio]{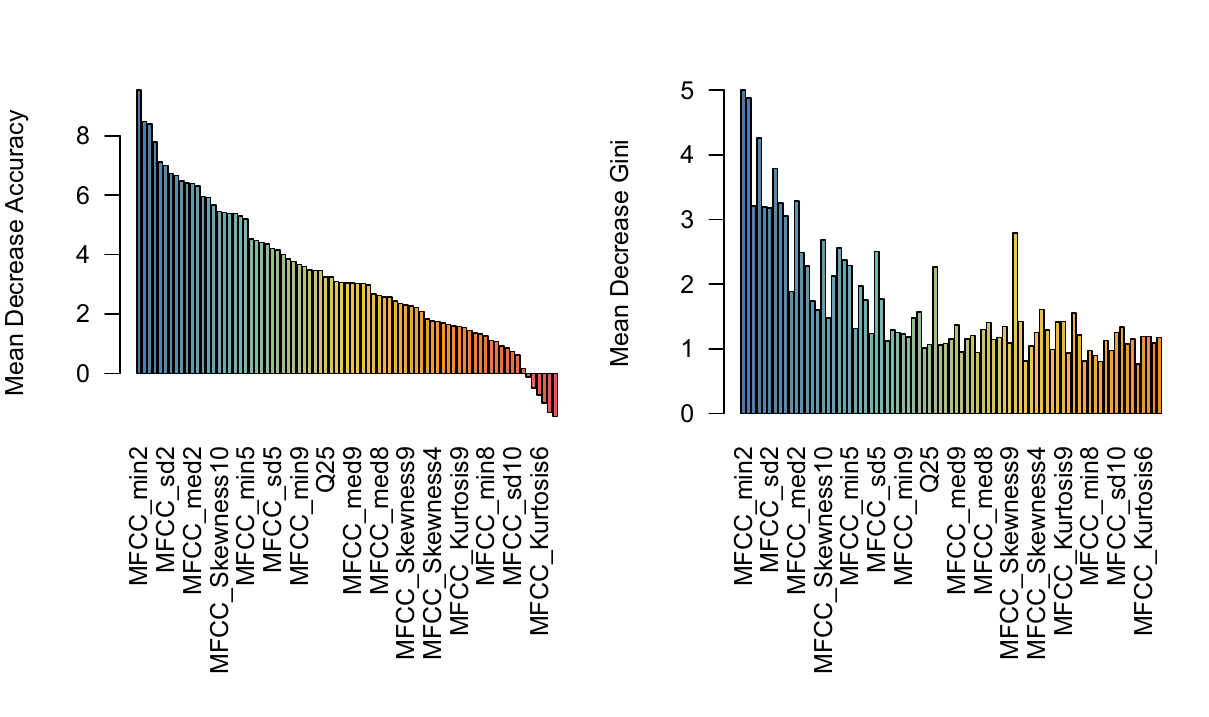}
    \caption{m=10 Data 1 - \textit{Passer domesticus} and \textit{Leiothrix lutea}}
    \label{m10}
\end{subfigure}

\caption{Importance variables by m for Data 1 - \textit{Passer domesticus} and \textit{Leiothrix lutea}.}\label{d1m}
\end{figure}

\noindent The variable importance analysis for the m=4 (Figure \ref{m4}) configuration, the ranking structure changes moderately, although Entropy and ZCR remain among the most influential variables. The reduced dimensional configuration appears to concentrate predictive information into a smaller subset of cepstral descriptors, particularly variables associated with medians, means, and maxima of lower-order \texttt{MFCC} coefficients, such as \texttt{MFCC\_med1}, \texttt{MFCC\_max1}, and \texttt{MFCC\_mean4}. Interestingly, quantile-based acoustic measures such as Q75 gain relative importance in this scenario, suggesting that upper-tail energy behavior becomes more informative under lower-dimensional representations. This may indicate that coarse spectral energy patterns are more robust than fine cepstral variations when the representation complexity is reduced.\\

\noindent The variable importance analysis for the m=10 (Figure \ref{m10}) configuration reveals that Entropy and Zero Crossing Rate (ZCR) are the most influential predictors according to both Mean Decrease Accuracy and Mean Decrease Gini criteria. Their dominant contribution suggests that spectral complexity and temporal signal irregularity are fundamental for discriminating between the vocalizations of \textit{Euphonia violacea} and \textit{Leiothrix lutea}. In particular, Entropy captures the distributional complexity of the acoustic spectrum, whereas ZCR reflects rapid temporal oscillations associated with fine-grained vocal structures.\\

\noindent Several MFCC-derived descriptors also exhibited substantial predictive relevance, especially lower-order coefficients and higher-order distributional moments such as skewness and kurtosis. Variables including \texttt{MFCC\_min6}, \texttt{MFCC\_Skewness7}, \texttt{MFCC\_med2}, and \texttt{MFCC\_Kurtosis8} indicate that both central tendency and asymmetry of cepstral representations contribute meaningfully to class separation. The simultaneous relevance of skewness and kurtosis metrics suggests that non-Gaussian properties of the cepstral distributions play an important role in distinguishing species-specific acoustic signatures.

\begin{figure}[H]
\centering

\begin{subfigure}{0.85\textwidth}
    \centering
\includegraphics[width=11cm,height=10.8cm,keepaspectratio]{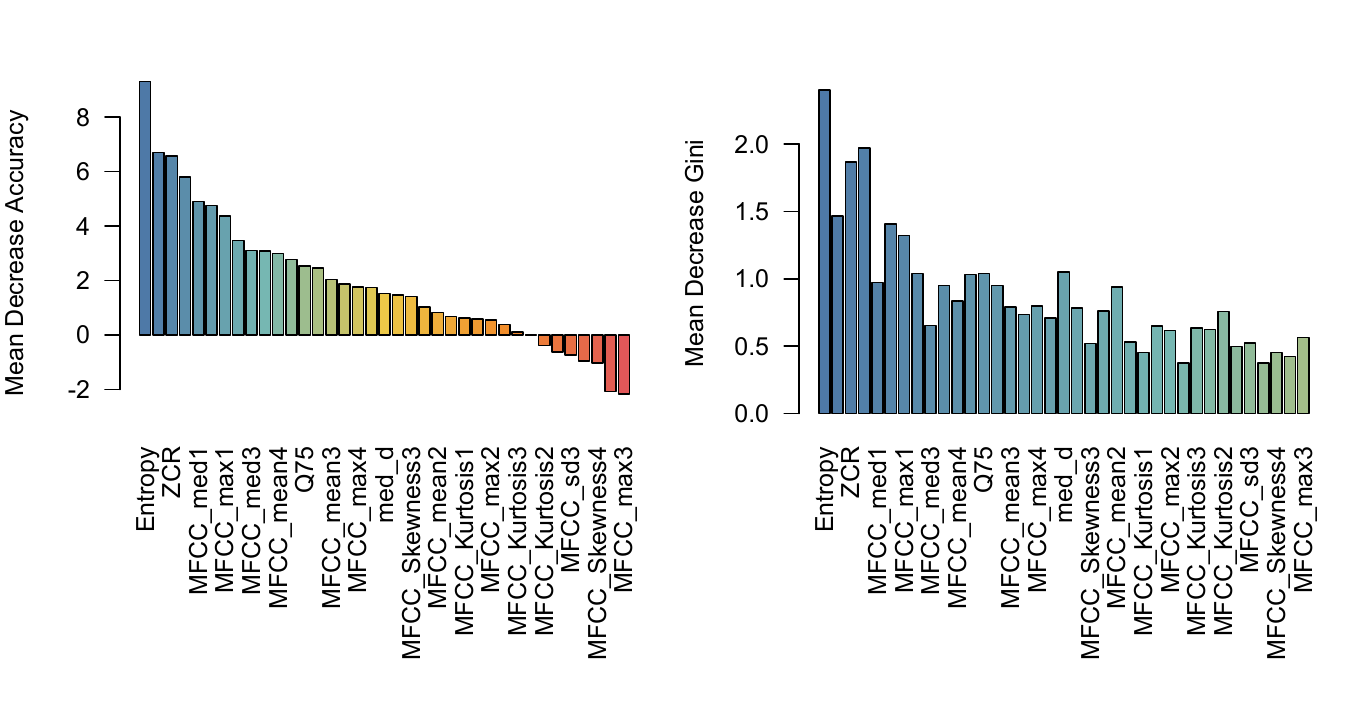}
    \caption{m=4 Data 2 - \textit{Euphonia violacea} and \textit{Leiothrix lutea}}
    \label{m4d2}
\end{subfigure}\hspace{0.2\textwidth} 
\begin{subfigure}{0.85\textwidth}
    \centering
\includegraphics[width=11cm,height=11.8cm,keepaspectratio]{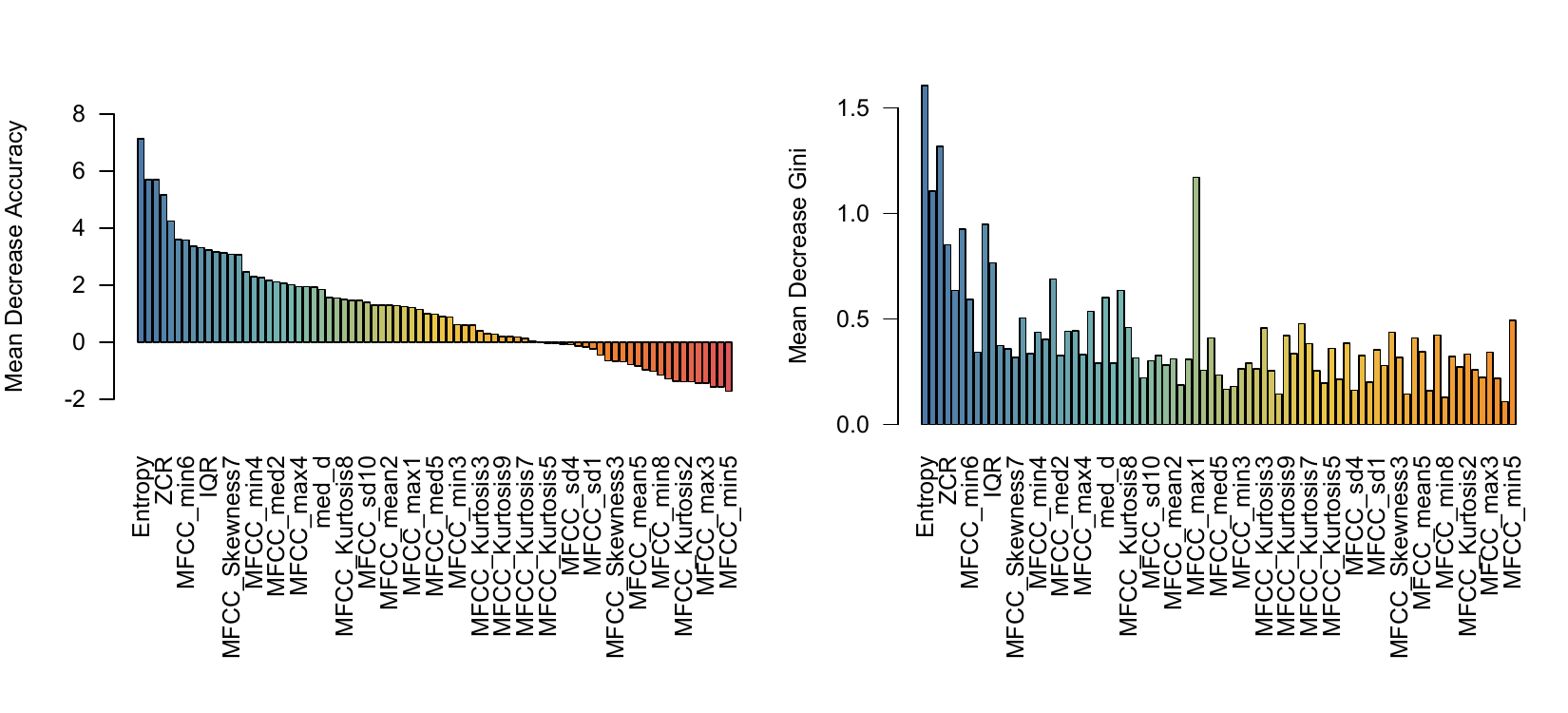}
    \caption{m=10 Data 2 - \textit{Euphonia violacea} and \textit{Leiothrix lutea}}
    \label{m10d2}
\end{subfigure}

\caption{Importance variables by m for Data 2 - \textit{Euphonia violacea} and \textit{Leiothrix lutea}.}
\end{figure}

\noindent In the m=4 configuration (Figure \ref{m4d2}), the importance ranking becomes even more concentrated around a small subset of \texttt{MFCC} descriptors, with \texttt{MFCC\_min2}, \texttt{MFCC\_mean2}, \texttt{MFCC\_med1}, and \texttt{MFCC\_max2} emerging as the most influential predictors. This pattern suggests that lower-order cepstral components retain the majority of discriminative information under reduced dimensionality. Additionally, spectral-energy measures such as Centroid and RMS become comparatively more relevant in this configuration, indicating that global frequency distribution and signal energy partially compensate for the reduction in cepstral detail. The inclusion of quantile measures (Q25 and Q75) further suggests that dispersion characteristics of the spectral distribution contribute to species differentiation.\\

\noindent The m=10 model for Data 2 (Figure \ref{m10d2}) demonstrates a distinct importance structure dominated primarily by \texttt{MFCC}-based descriptors rather than global acoustic statistics. Variables such as \texttt{MFCC\_min2}, \texttt{MFCC\_sd2}, \texttt{MFCC\_med2}, and \texttt{MFCC\_Skewness10} exhibit the highest predictive contributions, indicating that fine-scale cepstral variability is central to distinguishing the vocal emissions of \textit{Passer domesticus} and \textit{Leiothrix lutea}. The prominence of variance and skewness related \texttt{MFCC} measures suggests that temporal instability and asymmetry in spectral envelopes are particularly informative for this species pair. Additionally, the importance of lower-tail statistics (\texttt{MFCC\_min2}, \texttt{MFCC\_min5}, \texttt{MFCC\_min9}) indicates that low-energy spectral components contribute significantly to classification, potentially reflecting subtle harmonic or timbral differences between species.\\

\noindent Unlike Data 1 (Figure \ref{d1m}), traditional global descriptors such as Entropy and ZCR do not dominate the ranking, implying that species discrimination in this dataset relies more heavily on detailed cepstral structure than on broad spectral complexity measures. This difference may reflect intrinsic acoustic similarities between the species, requiring higher-resolution cepstral information to achieve adequate discrimination.

\noindent Table \ref{tab:importance_comparison} reveals a clear dominance of MFCC-derived descriptors in discriminating between Passer domesticus and Leiothrix lutea across both model configurations ($m=4$ and $m=10$). In particular, the variables associated with the lower-order MFCC coefficients, especially the minimum, mean, and median summaries of the first and second coefficients (e.g., \texttt{MFCC\_min2}, \texttt{MFCC\_mean1}, \texttt{MFCC\_mean2}, and \texttt{MFCC\_med2}), consistently exhibited the highest values for both Mean Decrease Accuracy (MDA) and Mean Decrease Gini (MDG). This pattern indicates that spectral envelope information contained in the low-frequency cepstral structure is the primary source of discrimination between the two species.

\noindent For the $m=4$ model, \texttt{MFCC\_min2} emerged as the most influential predictor, presenting the largest MDA (14.56) and MDG (9.14) values in the entire analysis. Similarly, \texttt{MFCC\_mean1}, \texttt{MFCC\_mean2}, \texttt{MFCC\_med1}, and \texttt{MFCC\_med2} also displayed remarkably high importance values, suggesting that central tendency measures of the first cepstral coefficients capture highly species-specific acoustic signatures. The prominence of these variables indicates that the distinction between the vocalizations of the two species is strongly associated with stable spectral patterns rather than with higher-order spectral fluctuations.\\

\noindent When increasing the dimensionality to $m=10$, the overall structure of variable importance remained stable, although the importance magnitudes generally decreased. This reduction is expected because the inclusion of additional MFCC dimensions distributes explanatory power across a larger number of correlated predictors. Nevertheless, the same subset of variables remained dominant, particularly \texttt{MFCC\_min2} (MDA = 9.54; MDG = 5.00), \texttt{MFCC\_mean1}, \texttt{MFCC\_mean2}, \texttt{MFCC\_med5}, and \texttt{MFCC\_med1}. The persistence of these variables across model specifications demonstrates the robustness of the acoustic information extracted from the lower cepstral components.\\

\noindent An additional important finding is that several higher-order MFCC descriptors introduced in the $m=10$ configuration exhibited low or even negative MDA values (e.g., \texttt{MFCC\_Kurtosis7}, \texttt{MFCC\_Skewness8}, and \texttt{MFCC\_max9}). Negative MDA values suggest that these variables do not contribute meaningfully to classification performance and may introduce noise or redundancy into the model. This result reinforces the notion that increasing the number of cepstral dimensions does not necessarily improve discrimination accuracy and may instead reduce interpretability by incorporating acoustically irrelevant information.\\

\noindent Beyond MFCC derived descriptors, traditional temporal and spectral measures such as Centroid, RMS, ZCR, Q25, and Q75 showed moderate but consistent importance across both models. Although these variables contributed to classification, their importance values were substantially lower than those observed for the leading MFCC features, indicating that broad spectral energy distribution and signal dispersion characteristics play a secondary role in species differentiation relative to cepstral structure.

\begin{table}[H]
\centering
\caption{Comparison of variable importance measures by Data 1 - \textit{Passer domesticus} and \textit{Leiothrix lutea} for $m=4$ and $m=10$. Blank cells indicate that the variable was not selected in the corresponding model.}
\label{tab:importance_comparison}
\tiny
\begin{tabular}{lrrrr}
\hline
\multirow{2}{*}{Variable} & \multicolumn{2}{c}{$m=4$} & \multicolumn{2}{c}{$m=10$} \\
\cline{2-5}
 & Mean Decrease Accuracy & Mean Decrease Gini & Mean Decrease Accuracy & Mean Decrease Gini \\
\hline
Centroid & 6.55 & 3.64 & 5.96 & 2.28 \\
Entropy & 2.44 & 3.34 & 1.36 & 1.22 \\
IQR & 3.55 & 2.03 & 3.60 & 1.18 \\
MFCC\_Kurtosis1 & 1.52 & 2.80 & 0.62 & 1.34 \\
MFCC\_Kurtosis10 &  &  & 1.58 & 1.41 \\
MFCC\_Kurtosis2 & 8.81 & 3.71 & 6.31 & 2.49 \\
MFCC\_Kurtosis3 & 0.69 & 1.62 & 1.33 & 0.81 \\
MFCC\_Kurtosis4 & 7.04 & 4.47 & 4.48 & 1.97 \\
MFCC\_Kurtosis5 &  &  & 1.07 & 0.81 \\
MFCC\_Kurtosis6 &  &  & -0.73 & 1.19 \\
MFCC\_Kurtosis7 &  &  & -1.32 & 1.09 \\
MFCC\_Kurtosis8 &  &  & 4.40 & 1.76 \\
MFCC\_Kurtosis9 &  &  & 1.57 & 1.42 \\
MFCC\_Skewness1 & 5.01 & 3.00 & 5.66 & 1.60 \\
MFCC\_Skewness10 &  &  & 5.46 & 2.68 \\
MFCC\_Skewness2 & 3.15 & 2.67 & 2.43 & 1.17 \\
MFCC\_Skewness3 & 1.52 & 1.77 & -0.49 & 0.76 \\
MFCC\_Skewness4 & 3.28 & 3.42 & 1.75 & 1.25 \\
MFCC\_Skewness5 &  &  & 1.10 & 0.90 \\
MFCC\_Skewness6 &  &  & 1.74 & 1.61 \\
MFCC\_Skewness7 &  &  & 1.65 & 0.99 \\
MFCC\_Skewness8 &  &  & -1.46 & 1.17 \\
MFCC\_Skewness9 &  &  & 2.30 & 1.09 \\
MFCC\_max1 & 7.02 & 3.53 & 6.42 & 1.88 \\
MFCC\_max10 &  &  & 2.27 & 2.79 \\
MFCC\_max2 & 8.52 & 4.28 & 4.16 & 1.77 \\
MFCC\_max3 & 3.38 & 2.46 & 3.46 & 1.01 \\
MFCC\_max4 & 6.77 & 3.80 & 5.92 & 1.74 \\
MFCC\_max5 &  &  & 4.53 & 1.31 \\
MFCC\_max6 &  &  & 2.97 & 1.20 \\
MFCC\_max7 &  &  & 3.46 & 1.57 \\
MFCC\_max8 &  &  & 2.57 & 1.14 \\
MFCC\_max9 &  &  & -1.00 & 1.19 \\
MFCC\_mean1 & 11.18 & 7.19 & 8.47 & 4.88 \\
MFCC\_mean10 &  &  & 3.24 & 2.27 \\
MFCC\_mean2 & 11.38 & 7.14 & 7.79 & 4.26 \\
MFCC\_mean3 & 2.80 & 2.21 & 1.54 & 0.93 \\
MFCC\_mean4 & 5.02 & 3.34 & 3.85 & 1.29 \\
MFCC\_mean5 &  &  & 6.67 & 3.25 \\
MFCC\_mean6 &  &  & 6.48 & 3.05 \\
MFCC\_mean7 &  &  & 4.36 & 1.24 \\
MFCC\_mean8 &  &  & 3.05 & 1.15 \\
MFCC\_mean9 &  &  & 2.21 & 1.42 \\
MFCC\_med1 & 10.02 & 6.99 & 6.74 & 3.79 \\
MFCC\_med10 &  &  & 5.38 & 2.12 \\
MFCC\_med2 & 11.52 & 5.89 & 6.40 & 3.28 \\
MFCC\_med3 & 5.96 & 2.63 & 3.06 & 1.08 \\
MFCC\_med4 & 7.09 & 3.76 & 3.47 & 1.47 \\
MFCC\_med5 &  &  & 8.39 & 3.21 \\
MFCC\_med6 &  &  & 5.30 & 2.37 \\
MFCC\_med7 &  &  & 4.00 & 1.12 \\
MFCC\_med8 &  &  & 2.62 & 1.30 \\
MFCC\_med9 &  &  & 3.04 & 1.37 \\
MFCC\_min1 & 8.40 & 5.20 & 7.11 & 3.19 \\
MFCC\_min10 &  &  & 3.02 & 0.95 \\
MFCC\_min2 & 14.56 & 9.14 & 9.54 & 5.00 \\
MFCC\_min3 & 3.44 & 2.31 & 0.93 & 1.13 \\
MFCC\_min4 & 6.60 & 3.30 & 2.58 & 1.41 \\
MFCC\_min5 &  &  & 5.21 & 2.29 \\
MFCC\_min6 &  &  & 3.02 & 1.15 \\
MFCC\_min7 &  &  & 2.66 & 0.94 \\
MFCC\_min8 &  &  & 1.25 & 0.97 \\
MFCC\_min9 &  &  & 3.66 & 1.23 \\
MFCC\_sd1 & 4.87 & 3.97 & 1.44 & 1.55 \\
MFCC\_sd10 &  &  & 0.74 & 1.25 \\
MFCC\_sd2 & 7.79 & 5.46 & 7.00 & 3.18 \\
MFCC\_sd3 & 4.22 & 2.19 & 0.85 & 0.98 \\
MFCC\_sd4 & 1.15 & 2.35 & 0.16 & 1.07 \\
MFCC\_sd5 &  &  & 4.21 & 2.50 \\
MFCC\_sd6 &  &  & 2.35 & 1.34 \\
MFCC\_sd7 &  &  & 3.10 & 1.06 \\
MFCC\_sd8 &  &  & -0.13 & 1.15 \\
MFCC\_sd9 &  &  & 2.08 & 0.81 \\
Q25 & 3.88 & 2.01 & 3.25 & 1.07 \\
Q75 & 4.43 & 1.94 & 3.77 & 1.25 \\
RMS & 5.71 & 3.55 & 5.38 & 2.56 \\
ZCR & 4.92 & 3.13 & 5.41 & 1.48 \\
med\_d & 4.71 & 2.00 & 1.70 & 1.29 \\
sd\_d & 2.39 & 2.18 & 1.83 & 1.05 \\
\hline
\end{tabular}
\label{tab:importance_m4_m10}
\end{table}

\noindent Table \ref{tab:importance_m4_m10_el} demonstrates a markedly different variable importance structure for the classification between Euphonia violacea and Leiothrix lutea when compared with the previous dataset. In this case, the discriminatory information was less concentrated in a small subset of MFCC descriptors and instead distributed across both classical acoustic indices and cepstral-based variables. This pattern suggests a more acoustically complex separation between the two species, where spectral energy distribution and temporal variability contribute jointly to classification performance.\\

\noindent For the $m=4$ configuration, Entropy emerged as the most influential non-MFCC variable, presenting a notably high Mean Decrease Accuracy (MDA = 9.31), followed by ZCR (6.57), \texttt{MFCC\_min1} (6.71), and Centroid (5.81). The strong importance of Entropy indicates that differences in spectral disorder or acoustic complexity play a central role in distinguishing the vocalizations of the two species. Likewise, the relevance of ZCR and Centroid suggests that temporal signal transitions and spectral balance are highly informative descriptors for this classification problem. These findings indicate that, unlike the previous dataset, the separation between species is not dominated exclusively by low-order cepstral structure.\\

\noindent Among the MFCC derived predictors, lower-order coefficients again showed greater importance than higher-order coefficients, although their magnitudes were considerably smaller than those observed in the \textit{Passer domesticus} versus \textit{Leiothrix lutea} comparison. Variables such as \texttt{MFCC\_mean1}, \texttt{MFCC\_med1}, \texttt{MFCC\_min1}, and \texttt{MFCC\_min4} consistently displayed moderate relevance, indicating that cepstral summaries still capture meaningful species-specific information. However, the relatively lower MDA and MDG values suggest weaker spectral separability between these species.\\

\noindent When the number of cepstral coefficients increased to $m=10$, the general importance pattern remained stable, but the explanatory contribution became even more diffuse across predictors. Entropy (MDA = 7.14), Centroid (5.71), ZCR (5.17), and \texttt{MFCC\_min1} (5.70) continued to rank among the most influential variables, demonstrating the robustness of these descriptors across model configurations. Additionally, some higher-order MFCC variables, such as \texttt{MFCC\_max6}, \texttt{MFCC\_min6}, and \texttt{MFCC\_Kurtosis10}, gained moderate relevance in the $m=10$ model, suggesting that finer spectral details may provide complementary discriminatory information when a larger cepstral representation is considered.\\

\noindent A particularly important result is the high frequency of negative or near-zero MDA values among several MFCC skewness, kurtosis, and standard deviation descriptors. Variables such as \texttt{MFCC\_Skewness2}, \texttt{MFCC\_max3}, \texttt{MFCC\_min5}, and \texttt{MFCC\_sd8} negatively affected predictive performance, indicating the presence of redundant or noisy information. This behavior reinforces the importance of variable selection procedures in high-dimensional acoustic models, especially when additional cepstral coefficients are incorporated.\\

\noindent Furthermore, the Mean Decrease Gini values were generally low across both configurations, rarely exceeding 2.5, which contrasts strongly with the previous dataset where several variables showed MDG values above 5. This finding suggests that the decision boundaries separating \textit{Euphonia violacea} and \textit{Leiothrix lutea} are less sharply defined, potentially reflecting greater overlap in their acoustic characteristics. Consequently, classification in this dataset appears to rely on the combined contribution of multiple moderately informative predictors rather than on a small set of highly dominant acoustic features.

\begin{table}[H]
\centering
\caption{Comparison of variable importance measures by Data 2 - \textit{Euphonia violacea} and \textit{Leiothrix lutea} for $m=4$ and $m=10$. Blank cells indicate that the variable was not selected in the corresponding model.}
\tiny
\begin{tabular}{lrrrr}
\hline
\multirow{2}{*}{Variable} & \multicolumn{2}{c}{$m=4$} & \multicolumn{2}{c}{$m=10$} \\
\cline{2-5}
 & Mean Decrease Accuracy & Mean Decrease Gini & Mean Decrease Accuracy & Mean Decrease Gini \\
\hline
Centroid & 5.81 & 1.97 & 5.71 & 1.11 \\
Entropy & 9.31 & 2.40 & 7.14 & 1.61 \\
IQR & 3.08 & 0.95 & 3.24 & 0.76 \\
MFCC\_Kurtosis1 & 0.63 & 0.45 & 0.04 & 0.26 \\
MFCC\_Kurtosis10 &  &  & 3.31 & 0.95 \\
MFCC\_Kurtosis2 & -0.38 & 0.76 & -1.39 & 0.33 \\
MFCC\_Kurtosis3 & 0.12 & 0.63 & 0.39 & 0.46 \\
MFCC\_Kurtosis4 & 1.46 & 0.78 & -0.08 & 0.39 \\
MFCC\_Kurtosis5 &  &  & -0.03 & 0.36 \\
MFCC\_Kurtosis6 &  &  & 1.15 & 0.26 \\
MFCC\_Kurtosis7 &  &  & 0.14 & 0.38 \\
MFCC\_Kurtosis8 &  &  & 1.50 & 0.46 \\
MFCC\_Kurtosis9 &  &  & 0.21 & 0.42 \\
MFCC\_Skewness1 & 1.75 & 0.71 & -0.14 & 0.33 \\
MFCC\_Skewness10 &  &  & -1.01 & 0.42 \\
MFCC\_Skewness2 & -2.08 & 0.42 & 0.30 & 0.25 \\
MFCC\_Skewness3 & 1.41 & 0.52 & -0.66 & 0.32 \\
MFCC\_Skewness4 & -1.02 & 0.45 & 0.28 & 0.14 \\
MFCC\_Skewness5 &  &  & -0.97 & 0.16 \\
MFCC\_Skewness6 &  &  & 0.21 & 0.33 \\
MFCC\_Skewness7 &  &  & 3.08 & 0.32 \\
MFCC\_Skewness8 &  &  & -1.43 & 0.22 \\
MFCC\_Skewness9 &  &  & -1.56 & 0.22 \\
MFCC\_max1 & 4.37 & 1.32 & 1.22 & 1.17 \\
MFCC\_max10 &  &  & -0.65 & 0.44 \\
MFCC\_max2 & 0.56 & 0.62 & -0.45 & 0.28 \\
MFCC\_max3 & -2.16 & 0.56 & -1.44 & 0.34 \\
MFCC\_max4 & 1.77 & 0.80 & 1.95 & 0.33 \\
MFCC\_max5 &  &  & -0.16 & 0.20 \\
MFCC\_max6 &  &  & 4.26 & 0.63 \\
MFCC\_max7 &  &  & 3.36 & 0.34 \\
MFCC\_max8 &  &  & -0.78 & 0.41 \\
MFCC\_max9 &  &  & 1.56 & 0.29 \\
MFCC\_mean1 & 4.76 & 1.41 & 3.60 & 0.93 \\
MFCC\_mean10 &  &  & 0.89 & 0.17 \\
MFCC\_mean2 & 0.82 & 0.94 & 1.29 & 0.31 \\
MFCC\_mean3 & 2.03 & 0.79 & 3.07 & 0.50 \\
MFCC\_mean4 & 2.99 & 0.83 & 3.14 & 0.36 \\
MFCC\_mean5 &  &  & -0.83 & 0.34 \\
MFCC\_mean6 &  &  & 1.31 & 0.33 \\
MFCC\_mean7 &  &  & 1.47 & 0.31 \\
MFCC\_mean8 &  &  & 1.30 & 0.28 \\
MFCC\_mean9 &  &  & 1.92 & 0.29 \\
MFCC\_med1 & 4.90 & 0.97 & 2.16 & 0.69 \\
MFCC\_med10 &  &  & -0.69 & 0.14 \\
MFCC\_med2 & 1.88 & 0.73 & 2.12 & 0.33 \\
MFCC\_med3 & 3.10 & 0.65 & 2.02 & 0.44 \\
MFCC\_med4 & 2.45 & 0.95 & 2.07 & 0.44 \\
MFCC\_med5 &  &  & 0.99 & 0.23 \\
MFCC\_med6 &  &  & 3.16 & 0.37 \\
MFCC\_med7 &  &  & 2.26 & 0.40 \\
MFCC\_med8 &  &  & 1.28 & 0.19 \\
MFCC\_med9 &  &  & 1.26 & 0.31 \\
MFCC\_min1 & 6.71 & 1.46 & 5.70 & 1.32 \\
MFCC\_min10 &  &  & -1.36 & 0.27 \\
MFCC\_min2 & 0.38 & 0.37 & -1.39 & 0.26 \\
MFCC\_min3 & 0.67 & 0.53 & 0.62 & 0.26 \\
MFCC\_min4 & 3.47 & 1.04 & 2.30 & 0.44 \\
MFCC\_min5 &  &  & -1.71 & 0.49 \\
MFCC\_min6 &  &  & 3.59 & 0.59 \\
MFCC\_min7 &  &  & 1.46 & 0.22 \\
MFCC\_min8 &  &  & -1.15 & 0.13 \\
MFCC\_min9 &  &  & 2.47 & 0.33 \\
MFCC\_sd1 & -0.01 & 0.62 & -0.23 & 0.35 \\
MFCC\_sd10 &  &  & 1.40 & 0.30 \\
MFCC\_sd2 & -0.63 & 0.50 & -1.29 & 0.32 \\
MFCC\_sd3 & -0.74 & 0.52 & -0.00 & 0.20 \\
MFCC\_sd4 & -0.96 & 0.38 & -0.08 & 0.16 \\
MFCC\_sd5 &  &  & 0.59 & 0.26 \\
MFCC\_sd6 &  &  & 1.00 & 0.41 \\
MFCC\_sd7 &  &  & -0.04 & 0.21 \\
MFCC\_sd8 &  &  & -1.57 & 0.11 \\
MFCC\_sd9 &  &  & 0.89 & 0.18 \\
Q25 & 1.03 & 0.76 & 1.95 & 0.54 \\
Q75 & 2.53 & 1.04 & 1.55 & 0.64 \\
RMS & 0.59 & 0.65 & 0.61 & 0.29 \\
ZCR & 6.57 & 1.87 & 5.17 & 0.85 \\
med\_d & 1.52 & 1.05 & 1.84 & 0.60 \\
sd\_d & 2.78 & 1.03 & 0.19 & 0.48 \\
\hline
\end{tabular}
\label{tab:importance_m4_m10_el}
\end{table}


\subsection{Selection of models}
\noindent The comparative evaluation of the classification models for Data 1 (\textit{Passer domesticus} and \textit{Leiothrix lutea} - Table \ref{tab:accuracy_comparison}) demonstrates substantial differences in predictive performance across algorithms and dimensional configurations (m=4 and m=10). The results indicate that increasing the dimensionality from m=4 to m=10 generally improved classification performance for the majority of models, particularly for Multinomial Logistic Regression and Support Vector Machines (SVM).\\

\noindent Among all evaluated methods, the SVM classifier achieved the best overall performance, yielding the highest Accuracy (0.85), Balanced Accuracy (0.84), and Cohen’s Kappa (0.69) under the m=10 configuration. These results suggest that the SVM was particularly effective in capturing the nonlinear decision boundaries associated with the acoustic structure of the species vocalizations. Moreover, the simultaneous attainment of high Sensitivity (0.91) and Specificity (0.77) indicates a well-balanced classification behavior, avoiding excessive bias toward either class. The relatively high Kappa coefficient further confirms substantial agreement beyond chance, reinforcing the robustness of the SVM model for birdsong discrimination. Taken together, these findings in Table \ref{tab:accuracy_comparison}, indicate that the m=10 representation generally provided superior discriminatory capacity, particularly for classifiers capable of exploiting complex multivariate relationships. The superior performance of SVM and Multinomial Logistic Regression highlights the importance of preserving richer acoustic information in the feature extraction stage, especially when dealing with highly variable bioacoustic signals.\\

\noindent The comparative performance of the supervised models evaluated on Data 2 - Table \ref{tab:accuracy_comparison_data2}, demonstrates that the Support Vector Machine (SVM) and Multinomial Logistic Regression models are the most robust frameworks for discriminating between \textit{Euphonia violacea} and \textit{Leiothrix lutea}. Under the $m = 10$ configuration, the SVM achieved the highest overall accuracy of 0.94 (Balanced Accuracy = 0.89), while the Multinomial Logistic Regression model yielded a highly competitive accuracy of 0.94 and the top Cohen's Kappa coefficient of 0.81. Moving from $m = 4$ to $m = 10$ systematically improved sensitivity across both classifiers, increasing from 0.72 to 0.86 for Logistic Regression and from 0.70 to 0.80 for SVM while preserving exceptional specificity levels above 0.95. This indicates that the addition of higher-order Mel-Frequency Cepstral Coefficients (MFCCs) captures crucial species-specific spectral nuances that are otherwise lost in lower-dimensional representations.\\

\noindent Conversely, other classifiers struggled with inherent structural limitations in this bioacoustic task. The Relevance Vector Machine (RVM) collapsed due to a severe class-prediction bias, resulting in a low accuracy of 0.29 with high sensitivity (0.97) but nearly non-existent specificity (0.10) at $m = 10$. This systemic imbalance was statistically validated by McNemar’s test, which rejected marginal homogeneity with absolute significance ($p < 0.001$). Additionally, the K-Nearest Neighbors (KNN) model proved highly conservative; while maintaining high specificity (>0.96), its sensitivity remained low (0.457 at $m = 10$), indicating high susceptibility to overlapping acoustic signatures in distance-based clustering. Crucially, the high p-values of McNemar’s test for both SVM ($p = 0.665$) and Logistic Regression ($p = 0.794$) at $m = 10$ confirm the statistical symmetry of their errors, demonstrating that these top-performing architectures provide unbiased and highly reliable predictions for ecological monitoring.

\begin{table}[H]
\centering
\caption{Comparison of classification results for m=4 and m=10 by Data 1 - \textit{Passer domesticus} and \textit{Leiothrix lutea} for 200 iterations.}
\label{tab:accuracy_comparison}
\tiny
\begin{tabular}{lcc|cc|cc|cc|cc}
\hline
Modelo & \multicolumn{2}{c|}{Random Forest} 
& \multicolumn{2}{c|}{RVM} 
& \multicolumn{2}{c|}{KNN} 
& 
\multicolumn{2}{c|}{Multinomial Logistic}
& \multicolumn{2}{c}{SVM} \\
\cline{2-11}
& \multicolumn{2}{c|}{m} &  \multicolumn{2}{c|}{m} &  \multicolumn{2}{c|}{m} &  \multicolumn{2}{c|}{m} &  \multicolumn{2}{c}{m} \\ 
\cmidrule(lr){2-3}\cmidrule(lr){4-5}\cmidrule(lr){6-7}\cmidrule(lr){8-9}\cmidrule(lr){10-11}
 & 4 & 10 & 4 & 10 & 4 & 10 & 4 & 10 & 4 & 10 \\ \hline

Accuracy 
& 0.7722 &  0.7800 &  
0.5585 & 0.5589 &
0.8009   & 0.7483 & 
0.7993 & 0.8466 & 
0.8418 & 0.8501 \\

Sensitivity 
& 0.7937  & 0.7931 &
0.9995 & 0.9994 & 
0.8988 & 0.8675  & 
0.8520  & 0.8958 &
0.9111 & 0.9151 \\

Specificity 
& 0.7437 & 0.7648  & 
0.0185  & 0.0196 & 
0.6744 & 0.5988 & 
0.7332 &  0.7860  &
0.7592   &   0.7725 \\

Balanced Accuracy & 0.7687  &  0.7789  & 
0.5090  & 0.5095  & 
0.7866 &  0.7331 & 
0.7926 &  0.8409 & 
0.8352 &  0.8438  \\

Kappa 
& 0.5347  &  0.5524 &
0.020 &  0.0210  & 
0.5837  & 0.4750 & 
0.5879  & 0.6853 & 
0.6752  &  0.6924 \\

Mcnemar's Test P-Value
& 0.6465  &  0.5858  & 
0.0000  & 0.0000 & 
0.2869 &  0.2215  &  
0.5904 & 0.5514 &  
0.3813 &  0.3993 \\

\hline
\end{tabular}
\end{table}
\begin{table}[H]
\centering
\caption{Comparison of classification results for m=4 and m=10 by Data 2 - \textit{Euphonia violacea} and \textit{Leiothrix lutea} for 200 iterations}
\label{tab:accuracy_comparison_data2}
\tiny
\begin{tabular}{lcc|cc|cc|cc|cc}
\hline
Modelo & \multicolumn{2}{c|}{Random Forest} 
& \multicolumn{2}{c|}{RVM} 
& \multicolumn{2}{c|}{KNN} 
& 
\multicolumn{2}{c|}{Multinomial Logistic}
& \multicolumn{2}{c}{SVM} \\
\cline{2-11}
& \multicolumn{2}{c|}{m} &  \multicolumn{2}{c|}{m} &  \multicolumn{2}{c|}{m} &  \multicolumn{2}{c|}{m} &  \multicolumn{2}{c}{m} \\ 
\cmidrule(lr){2-3}\cmidrule(lr){4-5}\cmidrule(lr){6-7}\cmidrule(lr){8-9}\cmidrule(lr){10-11}
 & 4 & 10 & 4 & 10 & 4 & 10 & 4 & 10 & 4 & 10 \\ \hline

Accuracy 
& 0.7255 & 0.7327 &
0.2955 & 0.2899  & 
0.8490 & 0.8509 &
0.9087& 0.9376 & 
0.9205 & 0.9398 \\

Sensitivity 
& 0.7501 & 0.7362 & 
0.9669  & 0.9682 &
0.3917& 0.4570 & 
0.7246  & 0.8651  & 
0.6965 & 0.8049 \\

Specificity 
& 0.7183  & 0.7319  & 
0.1118 &  0.1043  & 
0.9764 & 0.9608  & 
0.9596 & 0.9576  &
0.9824  & 0.9774  \\

Balanced Accuracy
& 0.7342 &  0.7340 & 
 0.5394&  0.5363 & 
 0.6840 & 0.7089 & 
 0.8421 &  0.9114 & 
 0.8394 & 0.8911 \\

Kappa 
& 0.3664 & 0.3753 & 
 0.0367 &  0.0336  &
 0.4392  & 0.4793 & 
 0.7116 & 0.8134 & 
 0.7376 & 0.8119 \\

Mcnemar's Test P-Value
&  0.1087  &  0.1521  & 
0.0000 & 0.0000  &
0.1411 & 0.2622 & 
0.6208 & 0.7941 &
 0.3958 & 0.6651 \\

\hline
\end{tabular}
\end{table}

\section{Final considerations}

\noindent The findings of this study underscore the critical importance of specialized preprocessing techniques in the analysis of bioacoustic signals from natural environments. The implementation of the bayesian wavelet shrinkage rule with an Epanechnikov prior proved to be a decisive factor in recovering the underlying bird songs from recordings with low signal to noise ratios (SNR). Unlike traditional thresholding methods, this approach provides an explicit, computationally efficient decision rule that is particularly robust under high noise levels, a common obstacle in field recordings.\\

\noindent Our comparative analysis o classification models reveals that the choice of both the algorithm and the feature space dimensionality is paramount. The Support Vector Machine (SVM) consistently outperformed other models, suggesting its superior ability to handle the complex, nonlinear decision boundaries inherent in species-specific acoustic signatures. Furthermore, the transition from a 4 to 10 dimensional feature space generally improved predictive metrics, highlighting that higher order cepstral information contains vital discriminatory details that simpler models might overlook.\\

\noindent The variable importance analysis provided significant ecological insights, showing that while global descriptors like spectral entropy and zero crossing rate are influential, the spectral envelope information captured by lower order MFCCs remains the primary source of discrimination between invasive species. This suggests that species identification relies heavily on stable, low-frequency cepstral structures.\\

\noindent The integrated pipeline presented here combining advanced Bayesian denoising with robust supervised learning offers a high performance solution for the automated monitoring of invasive birds
. Future research could explore the scalability of this framework to hyper-diverse soundscapes and the integration of deep learning architectures to further enhance the detection of rare or mimetic vocalizations in real-time ecological surveillance.

\subsection*{Acknowledgments}
\noindent Laura Lucia Dominguez Barrios was funded by the São Paulo Research Foundation (FAPESP), through grant No. 23/18444-4, linked to grant No. 22/04006-2 – Center for Molecular Plant Breeding, AP.PCPE. Fidel Aniano Causil Barrios received support from the Coordination for the Improvement of Higher Education Personnel – Brazil (CAPES) – Funding Code 001.\\

\bibliography{references}

\end{document}